\PassOptionsToPackage{dvipsnames}{xcolor}
\documentclass[10pt,twocolumn,letterpaper]{article}
\pdfoutput=1

\usepackage[margin=1in]{geometry}
\usepackage[many]{tcolorbox}
\usepackage{graphicx}
\usepackage{caption}
\usepackage{multicol}
\usepackage{amsmath}
\usepackage{mathptmx}
\usepackage{pgfplots}
\usepackage{multirow}
\usepackage{booktabs}
\usepackage{hhline}
\usepackage{subfigure} 
\usepackage{appendix}
\pgfplotsset{compat=1.18}

\definecolor{lightblue}{RGB}{230,240,255}
\definecolor{lightgreen}{RGB}{230,255,230}

\tcbset{
    enhanced,
    colback=lightblue,
    fonttitle=\bfseries,
    fontupper=\rmfamily
}

\usepackage{iccv}              
\usepackage{booktabs}
\usepackage{multirow}
\usepackage{colortbl}
\usepackage{pifont}     
\definecolor{darkgreen}{RGB}{50,100,0}
\definecolor{darkred}{RGB}{200, 0, 0}
\newcommand{\cmark}{\textcolor{darkgreen}{\ding{51}}} 
\newcommand{\xmark}{\textcolor{darkred}{\ding{55}}} 
\newcommand{\cxmark}{\ding{52}\rotatebox[origin=c]{-9.2}{\kern-0.7em\ding{55}}} 
\newcommand{\gray}[1]{\textcolor{gray}{#1}}

\definecolor{iccvblue}{rgb}{0.21,0.49,0.74}
\usepackage[pagebackref,breaklinks,colorlinks,allcolors=iccvblue]{hyperref}

\def\paperID{5735} 
\def\confName{ICCV}
\def\confYear{2025}

\title{MMCR: Benchmarking Cross-Source Reasoning in Scientific Papers}

\author{
Yang Tian$^{1}$\thanks{Co-first authors}
\qquad
Zheng Lu$^{1,2}$$^{*}$
\qquad
Mingqi Gao$^{1,3}$$^{*}$
\qquad
Zheng Liu$^{4}$
\qquad
Bo Zhao$^{1}$\thanks{Corresponding author}\\
\textsuperscript{\rm 1}  School of AI, Shanghai Jiao Tong University
\qquad
\textsuperscript{\rm 2} SCUT
\qquad 
\textsuperscript{\rm 3} SEU
\qquad
\textsuperscript{\rm 4} BAAI\\
\texttt{\{yangtian6781, bo.zhao\}@sjtu.edu.cn} \\
\url{https://github.com/yangtian6781/MMCR}
}
\begin{document}
\maketitle
\begin{abstract}
Fully comprehending scientific papers by machines reflects a high level of Artificial General Intelligence, requiring the ability to reason across fragmented and heterogeneous sources of information, presenting a complex and practically significant challenge. While Vision-Language Models (VLMs) have made remarkable strides in various tasks, particularly those involving reasoning with evidence source from single image or text page, their ability to use cross-source information for reasoning remains an open problem. This work presents MMCR, a high-difficulty benchmark designed to evaluate VLMs' capacity for reasoning with cross-source information from scientific papers. The benchmark comprises 276 high-quality questions, meticulously annotated by humans across 7 subjects and 10 task types. Experiments with 18 VLMs demonstrate that cross-source reasoning presents a substantial challenge for existing models. Notably, even the top-performing model, GPT-4o, achieved only 48.55\% overall accuracy, with only 20\% accuracy in multi-table comprehension tasks, while the second-best model, Qwen2.5-VL-72B, reached 39.86\% overall accuracy. Furthermore, we investigated the impact of the Chain-of-Thought (CoT) technique on cross-source reasoning and observed a detrimental effect on small models, whereas larger models demonstrated substantially enhanced performance. These results highlight the pressing need to develop VLMs capable of effectively utilizing cross-source information for reasoning.
\end{abstract}
\section{Introduction}

\begin{table*}[t]
\setlength{\tabcolsep}{3.5pt}
\centering
\small
\vspace{2ex}
\begin{tabular}{l|c|ccccccccc}
\toprule
\textbf{Benchmarks}  & \textbf{Avg. Pages} & \textbf{Cross-Page} & \textbf{Text} & \textbf{Chart} & \textbf{Table} & \textbf{Image} & \textbf{Pseudocode} & \textbf{Formula} & \textbf{CR} & \textbf{Annotatiom}\\
DocVQA~\cite{docvqa} & 1.0 & \xmark & \cmark & \cmark & \cmark & \cmark & \xmark & \xmark & \xmark & \cmark\\
ChartQA~\cite{chartqa} & 1.0 & \xmark & \xmark & \cmark & \xmark & \xmark & \xmark & \xmark & \xmark & \cxmark\\\
InfoVQA~\cite{chartqa} & 1.0 & \xmark & \xmark & \cmark & \cmark & \cmark & \xmark & \xmark & \xmark & \cmark\\
Charxiv~\cite{charxiv} & 1.0 & \xmark & \xmark & \cmark & \xmark & \xmark & \xmark & \xmark & \xmark & \cxmark\\
ArxivQA~\cite{arxivqa} & 1.0 & \xmark & \cmark & \cmark & \cmark & \cmark & \xmark & \xmark & \xmark & \xmark\\
MMSCI~\cite{mmsci} & 1.0 & \xmark & \cmark & \cmark & \cmark & \cmark & \xmark & \xmark & \xmark & \xmark\\
DUDE~\cite{dude} & 5.7 & \cmark & \cmark & \cmark & \cmark & \cmark & \xmark & \xmark & \xmark & \cmark\\
MP-DocVQA~\cite{mpdocvqa} & 8.3 & \xmark & \cmark & \cmark & \cmark & \cmark & \xmark & \xmark & \xmark & \cxmark\\
SlideVQA~\cite{slidevqa} & 20.0 & \cmark & \cmark & \cmark & \cmark & \cmark & \xmark & \xmark & \xmark & \cmark\\
MMLongBench-Doc~\cite{mmlongbench} & 47.5 & \cmark & \cmark & \cmark & \cmark & \cmark & \xmark & \xmark & \xmark & \cmark\\
\rowcolor[HTML]{F2F3F5}
MMCR & 19.0 & \cmark & \cmark & \cmark & \cmark & \cmark & \cmark & \cmark & \cmark & \cmark\\
\bottomrule
\end{tabular}
\caption{Comparison between our benchmark and previous related datasets. \textbf{Avg. Pages:} Average Pages. \textbf{Cross-Page:} Cross-Page Questions. \textbf{Text/Chart/Table/Image/Pseudocode/Formula:} Pure Text/Chart/Table/Image/Pseudocode/Formula Questions. \textbf{CR:} Cross-Source Reasoning Questions. \textbf{Annotation:} \cmark: Human Annotation, \xmark: Automatic Annotation, \cxmark: Semi-automatic Annotation.}
\label{tab:dataset_comparison}
\end{table*}

\begin{figure}
\centering
\includegraphics[width=0.5\textwidth]{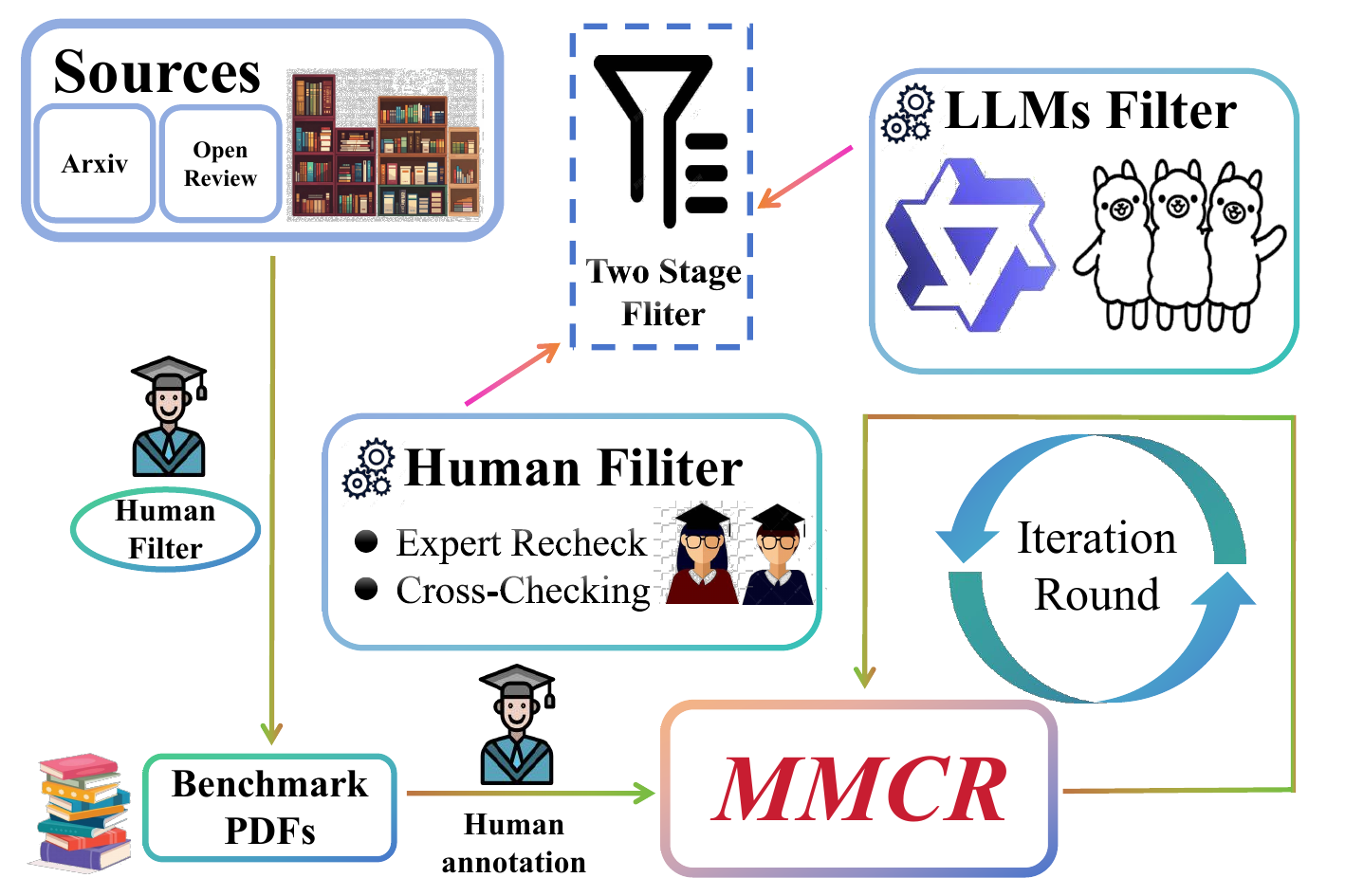}
\caption{The construction pipeline of MMCR.}
\label{pipline}
\end{figure}

Scientific papers encapsulate the advanced human intelligence, and it requires human experts years of study and practical experience to fully comprehend the content of these papers. Hence, comprehending scientific papers with machines is a long-standing but largely under-explored challenge. Scientific papers mainly contain elements that challenge a model's understanding and reasoning ability, such as mathematical formulas, pseudocode and figures with specialized knowledge and cross-source clues. 

Recently, a variety of VLMs have emerged, including proprietary ones such as GPT-4o~\cite{gpt4o}, Gemini-2.0~\cite{gemini}, and Claude-3~\cite{Claude}, as well as open-source ones such as InternVL2.5~\cite{internvl2.5}, Qwen2.5-VL~\cite{qwen2.5}, mPLUG-DocOwl2.0~\cite{mplug2}, LLaVA-OneVision~\cite{llavaonevision}, Molmo~\cite{molmo}, CogVLM2~\cite{cogvlm2}, and TextMonkey~\cite{textmonkey}. These models have demonstrated remarkable performance on general vision tasks, and many studies have explored their potential in document understanding, such as ChartQA~\cite{chartqa}, DocVQA~\cite{docvqa}, InfoVQA~\cite{infovqa}, DUDE~\cite{dude}, MP-DocVQA~\cite{mpdocvqa}, etc. However, the evaluation of VLMs' capabilities in scientific paper understanding is still relatively limited.

Evaluating models for understanding scientific papers holds significant practical importance. Previous research in this domain has presented three primary limitations: 1) \textbf{Dependence on Model-Generated Annotations:} The high degree of specialization in scientific papers makes manual annotation both time-consuming and costly. For instance, ArxivQA~\cite{arxivqa} uses GPT-4v to generate Question-Answer (QA) pairs, while SCI-CQA~\cite{scicqa} employs GPT-4o for the similar task. Although these generated QA pairs are manually reviewed, they still reflect biases inherent in the generative models, which could skew the evaluation of other models. 2) \textbf{Weak Relevance of Annotations to the Article’s Content:} The specialized nature of scientific papers presents a significant challenge in dataset construction. Charxiv~\cite{charxiv} collects papers from Arxiv but focuses solely on charts for real-world chart understanding, neglecting the specialized interpretation of these charts within the context of the articles. Similarly, MMSCI~\cite{mmsci}, which utilizes high-quality papers from \textit{Nature Communications}, concentrates on constructing datasets around figures, without addressing the content-related questions and answers. These efforts leverage the complexity of images in scientific papers but fail to tap into the specialized knowledge embedded within the scientific paper. 3) \textbf{Lack of Cross-Source Reasoning:} 
In scientific papers, cross-source interactions reveal insights that are absent from a single source. The ability to integrate and reason across diverse sources is crucial, as it demonstrates a deeper and more comprehensive understanding of the paper. However, the majority of existing research~\cite{charxiv, scicqa, arxivqa} primarily focuses on understanding individual source within scientific papers.
In summary, the absence of high-quality, professional benchmarks for scientific paper understanding has motivated the exploration of our work.

In this paper, we introduce MMCR, a benchmark specifically designed to evaluate the \textbf{M}ulti-\textbf{M}odality \textbf{C}ross-source \textbf{R}easoning capability of VLMs in scientific papers. Figure \ref{pipline} outlines the annotation pipeline of our benchmark. The benchmark has been meticulously curated by expert annotators, with each paper averaging 19 pages across 7 different subjects. We have crafted 10 distinct categories of questions, totaling 276 high-quality, domain-specific questions. The benchmark has undergone filtering assisted by large language models (LLMs) and rigorous manually cross-checking, which guarantees that each question is firmly tied to its cross-source clues and cannot be answered through alternative sources, thereby ensuring the robustness of the benchmark.

We conducted extensive experiments on MMCR to evaluate the cross-source reasoning capability of state-of-the-art VLMs in scientific papers. A total of 18 VLMs were evaluated, comprising 15 open-source models and 3 proprietary models. For each scientific paper, we re-render it into high-quality JPEG images and then are processed by the VLMs in an end-to-end approach. In other words, we expect VLMs to be able to comprehend scientific papers purely through visual input. Our results underscore the significant challenges these models face in performing cross-source reasoning of contents in scientific papers. Among the models tested, the best performer, GPT-4o, achieved an overall accuracy of only 48.55\%, while the second-best model, Qwen2.5-VL-72B, reached 39.86\%. Furthermore, we conducted a manual analysis of GPT-4o’s incorrect answers to identify the key bottlenecks in scientific paper understanding for current VLMs. The most frequent source of errors was perceptual error, which accounted for 27.5\% of the total, revealing the perceptual defects of current models when handling multi-page scientific papers with multiple information sources.
What's more, we explored the efficacy of CoT in cross-source reasoning and found that that its performance-enhancing benefits are predominantly evident in large-scale models, which reveals the limitations of CoT in enhancing cross-source reasoning for smaller models.
In summary, these findings highlight that cross-source reasoning in scientific papers remains a complex and ongoing challenge.

\section{Related Work}

\begin{figure*}
\centering
\includegraphics[width=0.8\textwidth]{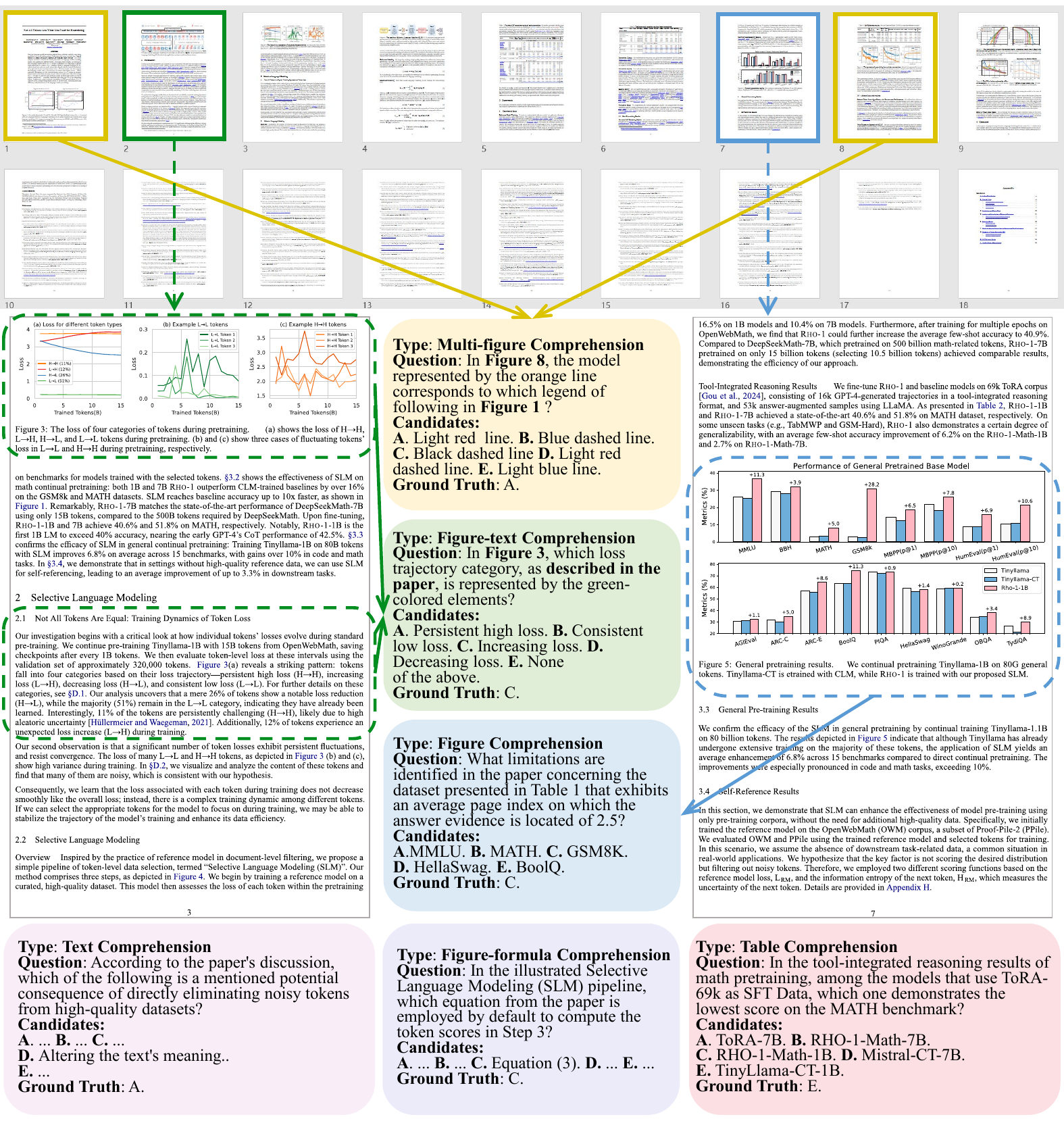}
\caption{Examples of MMCR. We meticulously curated 10 distinct question types, each accompanied by five answer options, to assess the ability of visual-language models to perform reasoning across multiple information sources.}
\label{main_figure}
\end{figure*}

\subsection{Scientific Paper Understanding}

Despite the increasing significance of automating the understanding of scientific papers, the datasets of scientific papers remain underdeveloped and insufficiently explored. In scientific paper understanding, data generation processes are mainly semi-automated. For example, Charxiv~\cite{charxiv} collects charts from Arxiv papers and employs GPT-4v to generate QA pairs, which are then manually reviewed to ensure data quality. ArxivQA~\cite{arxivqa} also utilizes GPT-4v for generating QA pairs specific to figures in Arxiv papers. MMSCI~\cite{mmsci}, despite relying on manual annotation and selecting high-quality figures from \texttt{Nature Communications}, does not generate content-related QA pairs. Instead, all tasks in the MMSCI datasets are based solely on captions, with minimal connection to the actual content of the papers. Moreover, none of these datasets address cross-page understanding of the entire paper, and the information sources required for answering the questions are singular. More detailed descriptions and comparisons are presented in Table \ref{tab:dataset_comparison}.

\subsection{Benchmarks for Document Understanding}

Compared to scientific paper understanding, document understanding is more advanced in its development. While the research of document understanding has shifted focus from single-page~\cite{chartqa,docvqa,infovqa} to multi-page understanding~\cite{mmlongbench, financebench, dude, mpdocvqa}, existing benchmarks for multi-page document understanding remain limited in critical aspects. MP-DocVQA~\cite{mpdocvqa}, an extension of the DocVQA~\cite{docvqa}, fails to adequately assess cross-page reasoning by omitting inter-page dependency questions. Although the DUDE~\cite{dude} dataset demonstrates rigorous annotation protocols with substantial manual effort, its average sample length of 5.7 pages per document provides insufficient scope for evaluating models on authentic multi-page challenges. SlideVQA~\cite{slidevqa} features cross-page questioning through 20-page slides, but the inherent low information density of slides makes its task less challenging. MMLongBench-Doc~\cite{mmlongbench} represents progress with comprehensive 47.5-page document samples across seven document types. However, it neglects to evaluate cross-source information synthesis, which is a crucial capability for real-world applications requiring integration of disparate information in a document. These limitations collectively highlight the need for benchmarks that simultaneously address extended document length, cross-page relationships, heterogeneous source integration, and domain-specific information density.

\subsection{Models for Document Understanding}

Document understanding models can be broadly categorized into two main approaches. The first relies on optical character recognition (OCR) tools, exemplified by models like LLaVA-Read~\cite{llavaread}, which use dual-stream vision encoder in conjunction with OCR tools to extract text from images, making it highly effective for text-rich image understanding tasks. The second approach is OCR-free, where models~\cite{textmonkey, qwen2vl, qwen2.5, Internvl2.0, internvl2.5} process documents end-to-end without the dependency of OCR tools. Currently, there is a clear trend in document understanding models evolving from OCR-dependent architectures to OCR-free approaches.

\begin{table}[t!]
    \centering
    \small
    \begin{tabular}{lc}
     \toprule
     \textbf{Statistic} & \textbf{Number} \\
     \midrule
      \textbf{Scientific Paper} & 31 \\
      ~- Avg./Max. Pages &  19 / 35 \\
      ~- Avg./Max. Number of Options & 5 / 5 \\
      ~- Average Size (px) & $1573 \times 1210$\\
      ~- Maximum Size (px) & $1584 \times 1224$\\
    \midrule
    \textbf{Subject} & 7 \\
     \midrule
      ~- 2D Object Recognition & 4 (12.90\%)\\
      ~- Efficient AI & 2 (6.45\%)\\
      ~- Multimodal Learning & 15 (48.39\%)\\
      ~- Datasets and Evaluation & 4 (12.90\%)\\
      ~- 3D from Multi-View and Sensors & 1 (3.23\%)\\
      ~- Large Language Model & 2 (6.45\%)\\
      ~- Robotics & 3 (9.68\%)\\
     \midrule
      \textbf{Total Questions} & 276 \\
  ~- Figure Comprehension & 43 (15.58\%) \\
  ~- Multi-Figure Comprehension & 20 (7.25\%)  \\
  ~- Figure-Table Comprehension & 26 (9.42\%) \\
  ~- Figure-Text Comprehension & 53 (19.20\%) \\
  ~- Figure-Formula Comprehension & 13 (4.71\%) \\
  ~- Table Comprehension & 62 (22.46\%) \\
  ~- Multi-Table Comprehension & 10 (3.62\%) \\
  ~- Text Comprehension & 35 (12.68\%) \\
  ~- Formula Comprehension & 9 (3.26\%) \\
  ~- Pseudocode Comprehension & 5 (1.81\%) \\
     \bottomrule
     \end{tabular}
     \vspace{2ex}
    \caption{Benchmark Statistics.}
    \label{statistic}
\end{table}
\section{MMCR Benchmark}

MMCR is a comprehensive and challenging benchmark, which is designed to evaluate the cross-source reasoning capability of VLMs using multiple intertwined information sources in scientific papers. In this section, we detail the processes of data collection, question and answer annotation, quality control, and evaluation method. Finally, we present the statistical data of MMCR.

\begin{table*}[t!]
\setlength{\tabcolsep}{2.8pt}
\small
\begin{center}
\begin{tabular}{lccccccccccccc}
\toprule
\textbf{Model} & \textbf{Date} & \textbf{\#Param} & \textbf{Overall} & \textbf{FIC} & \textbf{MF} & \textbf{FTA} & \textbf{FTE} & \textbf{FF} & \textbf{TAC} & \textbf{MT} & \textbf{TEC} & \textbf{FOC} & \textbf{PC} \\
\midrule
\multicolumn{14}{c}{\textit{Open-Source Multimodal Large Language Models}}\\
\midrule
{\gray{\textit{\textbf{4-9B Models}}}}\\
$\text{Phi-3.5-Vision~\cite{phi3}}$  
& 2024-09 & 4B& 4.71 & 6.98 & 5.00 & 0.00 & 5.66 & 0.00 & 3.23 & 0.00 & 11.43 & 0.00 & 0.00\\
$\text{YI-VL~\cite{yivl}}^{*}$  
& 2024-06 & 6B& 4.35 & 2.33 & 15.00 & 3.85 & 5.66 & 7.69 & 1.61 & 0.00 & 5.71 & 0.00 & 0.00\\
$\text{Molmo-7B-O~\cite{molmo}}^{*}$  
& 2024-09 & 7B& 0.36 & 0.00 & 0.00 & 0.38 & 0.00 & 0.00 & 0.00 & 0.00 & 0.00 & 0.00 & 0.00\\
$\text{XComposer2.5~\cite{xcomposer-2.5}}^{*}$  
& 2024-07 & 7B& 3.99 & 4.65 & 20.00 & 0.00 & 3.77 & 0.00 & 3.22 & 0.00 & 2.86 & 0.00 & 0.00\\
$\text{mPLUG-Owl3~\cite{mplug3}}$  
& 2024-07 & 7B& 7.61 & 18.60 & 20.00 & 3.85 & 3.77 & 7.69 & 4.84 & 0.00 & 5.71 & 0.00 & 0.00\\
$\text{Qwen2.5-VL~\cite{qwen2.5}}$  
& 2025-01 & 7B& 21.74 & 16.28 & 25.00 & 7.69 & 26.42 & 30.77 & 16.13 & 10.00 & 45.71 & 11.11 & 0.00\\
$\text{Bunny~\cite{bunny}}^{*}$  
& 2024-06 & 8B& 4.71 & 4.65 & 10.00 & 3.85 & 1.89 & 0.00 & 4.84 & 0.00 & 5.71 & 11.11 & 20.00\\
$\text{Idefics3~\cite{Idefics3}}$  
& 2024-08 & 8B& 9.78 & 11.63 & 10.00 & 7.69 & 7.55 & 15.38 & 6.45 & 10.00 & 20.00 & 0.00 & 0.00\\
$\text{MiniCPM-o 2.6~\cite{minicpm}}$
& 2025-01 & 8B& 10.51 & 13.53 & 5.00 & 11.54 & 5.66 & 0.00 & 6.45 & 0.00 & 31.43 & 0.00 & 20.00\\
$\text{InternVL2.5~\cite{internvl2.5}}$  
& 2024-12 & 8B& 12.32 & 11.63 & 15.00 & 7.69 & 13.21 & 7.69 & 4.84 & 0.00 & 31.43 & 22.22 & 0.00\\
$\text{GLM-4v~\cite{chatglm}}^{*}$  
& 2024-06 & 9B& 2.54 & 2.33 & 5.00 & 0.00 & 5.66 & 0.00 & 1.61 & 0.00 & 0.00 & 11.11 & 0.00\\
{\gray{\textit{\textbf{11-78B Models}}}}\\
$\text{Llama3.2-Vision~\cite{llama3.2}}^{*}$  
& 2024-09 & 11B& 1.81 & 0.00 & 0.00 & 0.00 & 0.00 & 7.69 & 3.23 & 0.00 & 5.71 & 0.00 & 0.00\\
$\text{DeepSeek-VL2~\cite{deepseekvl2}}$  
& 2024-05 & 27B& 1.09 & 0.00 & 0.00 & 0.00 & 5.66 & 0.00 & 0.00 & 0.00 & 0.00 & 0.00 & 0.00\\
$\text{InternVL2.5~\cite{internvl2.5}}$  
& 2024-12 & 78B& 14.49 & 18.60 & 20.00 & 7.69 & 11.32 & 7.69 & 6.45 & 0.00 & 34.29 & 22.22 & 20.00\\
$\text{Qwen2.5-VL~\cite{qwen2.5}}$  
& 2025-01 & 72B& \underline{39.86} & \underline{30.23} & \underline{35.00} & 26.92 & \underline{50.94} & 38.46 & \underline{35.48} & 0.00 & \textbf{68.57} & 33.33 & \underline{40.00}\\
\midrule
\multicolumn{14}{c}{\textit{Proprietary Multimodal Large Language Models}}\\ 
\midrule
$\text{GPT-4o mini~\cite{gpt4o}}$  
 & 2024-07 & - & 32.61 & 25.58 & \underline{35.00} & 26.92 & 39.62 & \underline{46.15} & 30.65 & 10.00 & 37.14 & 33.33 & \underline{40.00} \\
$\text{GeminiFlash2.0~\cite{gemini}}$  
 & 2024-12 & - & 34.42 & 23.26 & 30.00 & \underline{30.77} & 39.62 & 30.77 & \underline{35.48} & \textbf{30.00} & 40.00 & \underline{44.44} & \textbf{60.00} \\
 $\text{GPT-4o~\cite{gpt4o}}$  
 & 2024-11 & - & \textbf{48.55} & \textbf{44.19} & \textbf{40.00} & \textbf{46.15} & \textbf{54.72} & \textbf{84.61} & \textbf{37.10} & \underline{20.00} & \underline{65.71} & \textbf{55.56} & \underline{40.00} \\
\bottomrule
\end{tabular}
\vspace{2ex}
\caption{Evaluation of various models on MMCR. We evaluate model performance across ten categories of scientific comprehension tasks: Figure Comprehension (\textbf{FIC}), Multi-figure Comprehension (\textbf{MF}), Figure-Table Comprehension (\textbf{FTA}), Figure-Text Comprehension (\textbf{FTE}), Figure-Formula Comprehension (\textbf{FF}), Table Comprehension (\textbf{TAC}), Multi-table Comprehension (\textbf{MT}), Text Comprehension (\textbf{TEC}), Formula Comprehension (\textbf{FOC}), Pseudocode Comprehension (\textbf{PC}). For each category, we report both individual accuracy rates and overall performance metrics. The best results are marked \textbf{bold} and the second results are \underline{underlined}. ${*}$ indicate that the input images are concatenated into one image.} 
\label{main_tab}
\end{center}
\end{table*}

\subsection{Data Collection}

We collected scientific papers from both Arxiv and high-quality publications on the OpenReview platform. Due to the highly specialized nature of scientific papers, annotating them presents significant challenges for non-experts. To address this, we engaged expert annotators who are not only proficient in English reading and writing but also have extensive experience in their respective research fields. The collected papers were distributed in batches to the annotators, who were given the flexibility to select papers within their areas of expertise for annotation.

\subsection{Question and Answer Annotation}

Before beginning the formal annotation process, we first pre-define the annotation categories and conduct annotation exercises. These categories are based on the sources of information needed to answer the questions. Specifically, the annotation categories are as follows:

1) \textbf{Figure Comprehension:} Figures in scientific papers serve as visual representations of models, theories, data, and other key elements. Questions based on these figures assess the model's ability to interpret and understand the visual information they convey.
What's more, the annotated questions do not specify which figure is being referenced, thereby evaluating the model’s ability to comprehend figures in the context of a multi-page scientific paper.

2) \textbf{Multi-Figure Comprehension:} In scientific papers, figures are often not standalone elements but are interconnected with other figures. By reasoning across multiple figures, it avoids being limited to the interpretation of individual figures, leading to more comprehensive conclusions. These questions evaluate the model's ability to utilize information across figures to answer effectively. Through meticulous verification, we ensure that all the necessary evidence for reasoning is confined to the relevant set of figures, with no information sourced from elsewhere.

3) \textbf{Figure-Table Comprehension:} The figure provides rich graphical information, while the table offers detailed structured data. These questions require VLMs to grasp the underlying relationships between graphical and structured information and perform reasoning across both sources.

4) \textbf{Figure-Text Comprehension:} Figures in scientific papers enormously necessitate specialized contextual knowledge for full comprehension. Questions of this task evaluate a model's ability to integrate information from both figures and text content to formulate accurate answers. If the model lacks a deep understanding of the text or fails to interpret the figures properly, these questions can pose considerable challenges.

5) \textbf{Figure-Formula Comprehension:} Formulas provide essential supplementary details that are crucial for understanding the content of figures. These questions evaluate a model’s ability to answer queries that require interpreting both figures and formulas. To generate accurate responses, the model must effectively integrate the relevant formula with the corresponding figure.

6) \textbf{Table Comprehension:} Tables in scientific papers serve as a foundational component for empirical analysis. These questions are designed to evaluate the model’s ability to extract information from individual tables and reason without explicitly referencing a specific table in the queries. This approach assesses the model’s capacity to interpret and navigate tables within a scientific paper.

7) \textbf{Multi-table Comprehension:} Scientific papers frequently include multiple tables, and drawing conclusions through cross-table comparisons is a common practice. This task assesses a model's capability to extract and integrate information from several tables, enabling it to synthesize insights from diverse data sources effectively.

8) \textbf{Text Comprehension:} Given that scientific papers are text-dense, these questions evaluate a model’s ability to extract and understand meaningful information from scientific papers.

9) \textbf{Formula Comprehension:} Formulas present condensed, symbolic information, which poses a greater challenge for the model compared to plain text. This task tests the model’s ability to understand and extract information from the formulas in scientific papers.

10) \textbf{Pseudocode Comprehension:} Although pseudocode appears less frequently than figures or tables in scientific papers, it often provides crucial additional context. These questions test the model’s ability to interpret the symbolic and graphical components of pseudocode.

Figure \ref{main_figure} presents an annotated example. The questions are designed to assess the model’s ability not only to derive answers from cross-source information but also to comprehend content across pages, as relevant evidence may be distributed throughout the document. This presents heightened challenges for the model’s performance. Detailed examples are provided in the supplementary material.

In our answer annotation process, we employed a multiple-choice format that required annotators to generate a minimum of eight questions per scientific paper. To ensure diversity in question types, annotators were instructed to avoid repetitive question patterns. Each question was mandated to have five distinct answer options; questions that could not meet this requirement were excluded. Papers yielding fewer than eight valid questions were omitted from the annotation.

\subsection{Quality Control}

To enhance the annotation quality of the benchmark, we implemented a two-round semi-automatic quality control procedure that combines the merits of manual annotations and LLMs.

\textbf{LLMs-Assisted Data Filtering.} Despite the meticulous efforts of our annotators, some questions addressed general knowledge that large language models could accurately answer based solely on their internal knowledge or by reasoning directly from the question and options. To evaluate this, we employed two language models, Llama-3.3-70B~\cite{llama3.2} and Qwen2.5-72B~\cite{qwen2.5llm}, to perform reasoning using only the questions and options as input, without incorporating any associated scientific papers, images, or supplementary text. Questions that the models answered correctly were flagged as potential low-quality questions. However, before discarding these questions entirely, we carefully reviewed the reasoning processes of the models to ensure their correct answers were not the result of random guessing.

\textbf{Cross-Checking.} Annotators conducted parallel cross-checking of each other's work to enhance the quality of the benchmark. Errors identified during this process were either corrected or eliminated. Furthermore, a voting mechanism was implemented to filter out question-and-answer pairs that were overly similar in structure or content, ensuring greater diversity and reliability in the final benchmark.

\subsection{Evaluation Method} 

With only five answer options, randomly selecting an option would result in an accuracy of approximately 20\%, which potentially reduces the discernible performance differences among VLMs. To address this, and inspired by MMBench~\cite{mmbench}, we implemented a circular evaluation to further minimize the likelihood of the model guessing the correct answer. In this approach, the order of the answer options is rotated five times and presented to the VLMs. A model is considered to have successfully solved the problem only if it answers correctly in all five tests.

\subsection{Statistics of Benchmark}
The detailed statistics of our benchmark are presented in Table \ref{statistic}. This benchmark is distinguished by its focus on cross-source reasoning, encompassing a total of 31 scientific papers with an average length of 19 pages per paper, spanning 7 diverse subjects. A total of 276 questions have been annotated, distributed across 10 distinct categories, with an average of 8.9 questions per paper. This comprehensive benchmark is designed to rigorously evaluate the capability of VLMs to perform reasoning by integrating information from multiple sources within scientific papers.

\section{Experiments and Analysis}

\subsection{Evaluation Protocol}

We conduct a three-step evaluation protocol: response generation, answer extraction, and score calculation. Specifically, in the answer generation stage, we deliberately refrain from imposing task-specific constraints on the model, allowing it to generate responses freely. Subsequently, we apply a heuristic rule-based extraction method to extract the model's output. The specific heuristic rule-based extraction method is described in detail in the supplementary material. Finally, we compute the exact match score to evaluate the model's performance. We report both the overall accuracy as well as the accuracy for each individual category.

\subsection{Experimental Setup}

\begin{figure}
\centering
\includegraphics[width=0.5\textwidth]{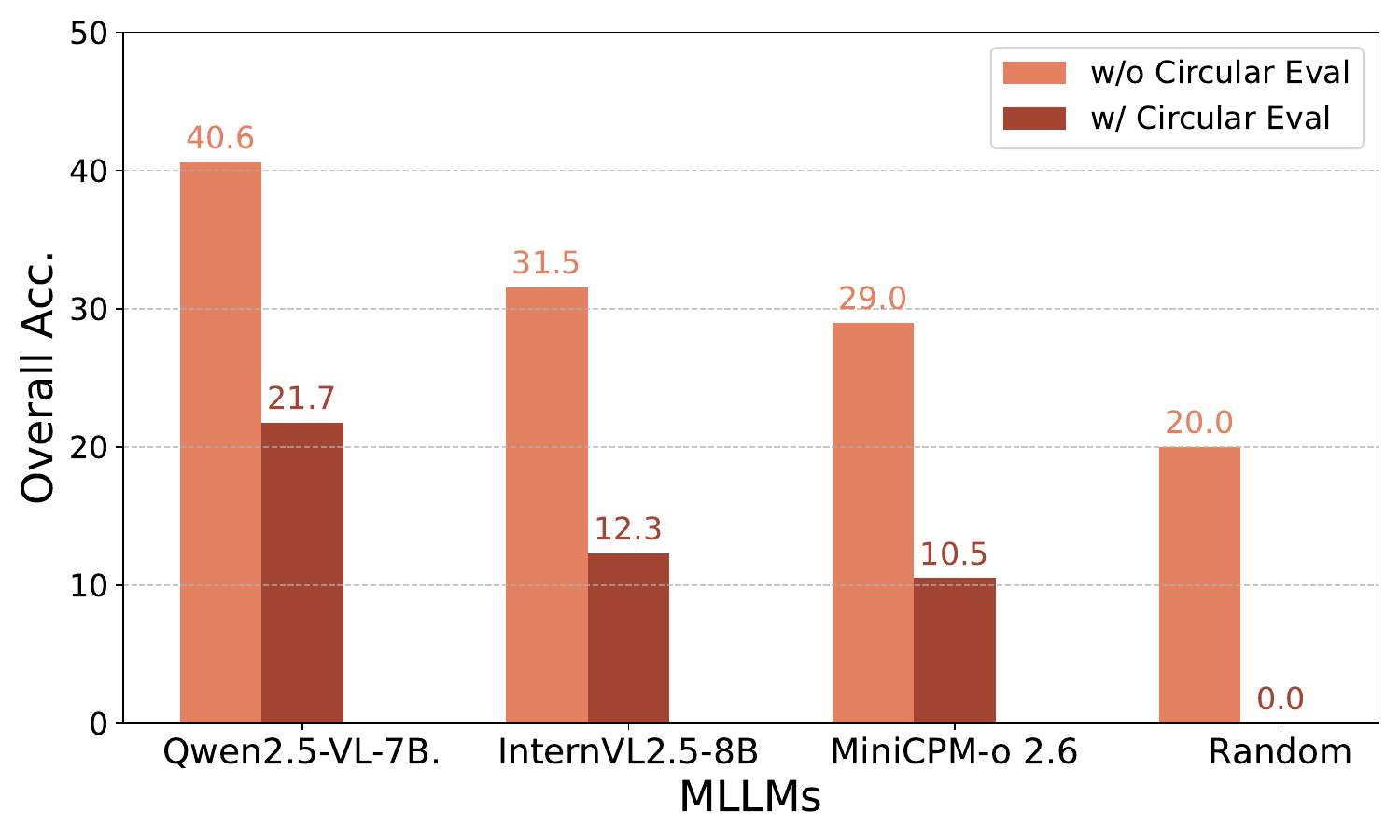}
\caption{Comparison of the overall accuracy of VLMs with and without circular eval.}
\label{circular_eval}
\end{figure}

We evaluated 18 VLMs, comprising 15 open-source models and 3 proprietary models. To rigorously assess the models' capability for reasoning across multiple sources of information in scientific papers, we re-rendered the PDF-format papers into high-resolution JPEG images at 144 DPI and fed them into the VLMs in an end-to-end approach. However, not all models support multi-image input. For those not trained on multi-image or video datasets, we concatenated the high-resolution JPEG images into a single large image before inputting it into the models. For models that do support multi-image input, we conducted evaluations using the native multi-image input method.

\subsection{Main Results}

\begin{figure}
\centering
\includegraphics[width=0.5\textwidth]{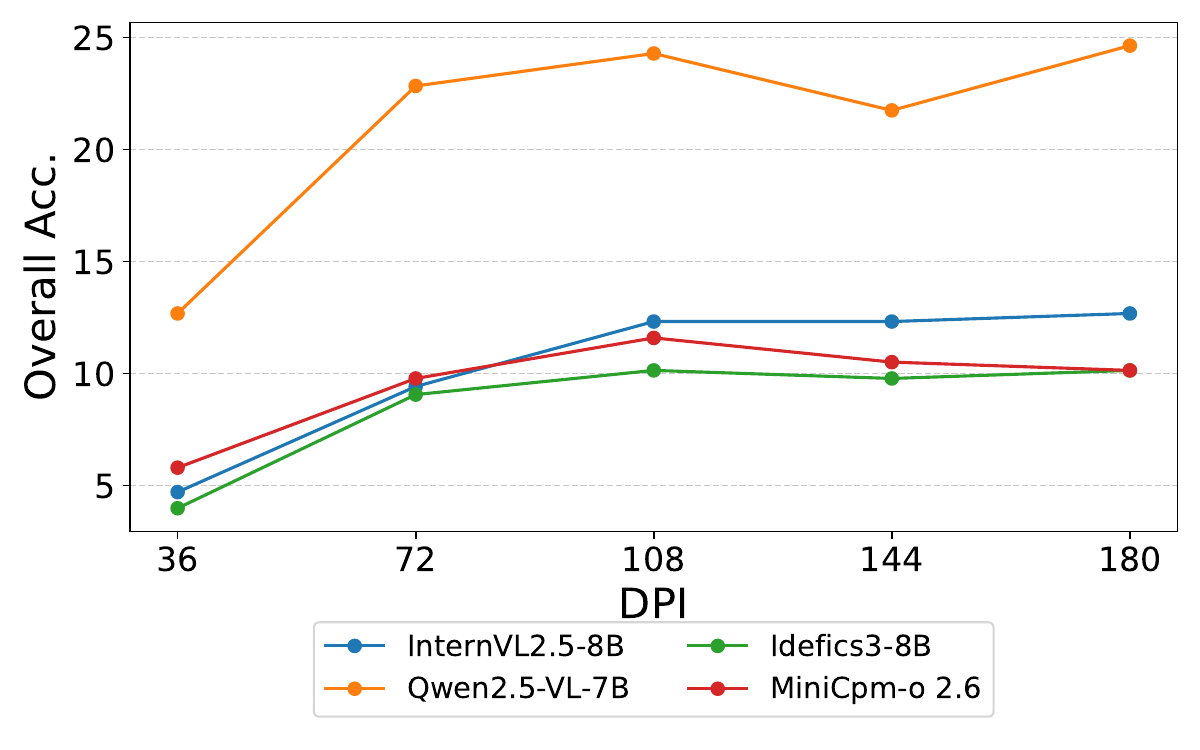}
\caption{Model performance on increasing rendering DPI.}
\label{resolution}
\end{figure}

The experimental results are comprehensively summarized in Table \ref{main_tab}, which details both category-specific and overall accuracy metrics. From these findings, three principal conclusions emerge:

\textbf{Proprietary models demonstrate superior performance.} The proprietary GPT-4o model achieved state-of-the-art results on our benchmark, attaining overall accuracy of 48.55\%. This represents an 8.69\% performance advantage over the leading open-source model, Qwen2.5-VL-72B. Notably, GPT-4o matched or exceeded Qwen2.5-VL-72B's performance across nine distinct question categories, underscoring the persistent performance gap in multi-modal reasoning capabilities between proprietary and open-source architectures for scientific document analysis.

\textbf{Current VLMs face significant challenges in reasoning with cross-source information.} Even the highest-performing proprietary VLM (GPT-4o) failed to surpass 50\% overall accuracy, while demonstrating sub-50\% performance in five question categories. Open-source models exhibited more pronounced limitations, with most models achieving less than 10\% overall accuracy. These results collectively highlight the inherent challenge of our benchmark.

\textbf{Tabular comprehension presents unique difficulties.} Comparative analysis revealed that 72.22\% (13/18) of evaluated models exhibited equal or inferior performance on table comprehension questions compared to figure comprehension questions. When extended to multi-element comprehension tasks, all tested models demonstrated equivalent or reduced accuracy. This systematic performance discrepancy suggests inherent challenges in tabular data interpretation within complex scientific papers.


\subsection{Is Circular Evaluation more Robust?}

In Figure \ref{circular_eval}, we evaluated three models and theoretical random selection with and without circular evaluation. As shown in the figure, all three models experienced significant performance degradation under circular evaluation, with MiniCPM-o 2.6's performance dropping to approximately 36\% of its performance without circular evaluation. The accuracy of random selection decreased from 20\% to ${0.2^{5} \times 100\%}$. These results demonstrate that circular evaluation substantially reduces the probability of obtaining correct answers through random selection, validating the robustness of our benchmark.

\subsection{Does High-Resolution Image Input Improve Model Performance?}

\begin{figure}
\centering
\includegraphics[width=0.5\textwidth]{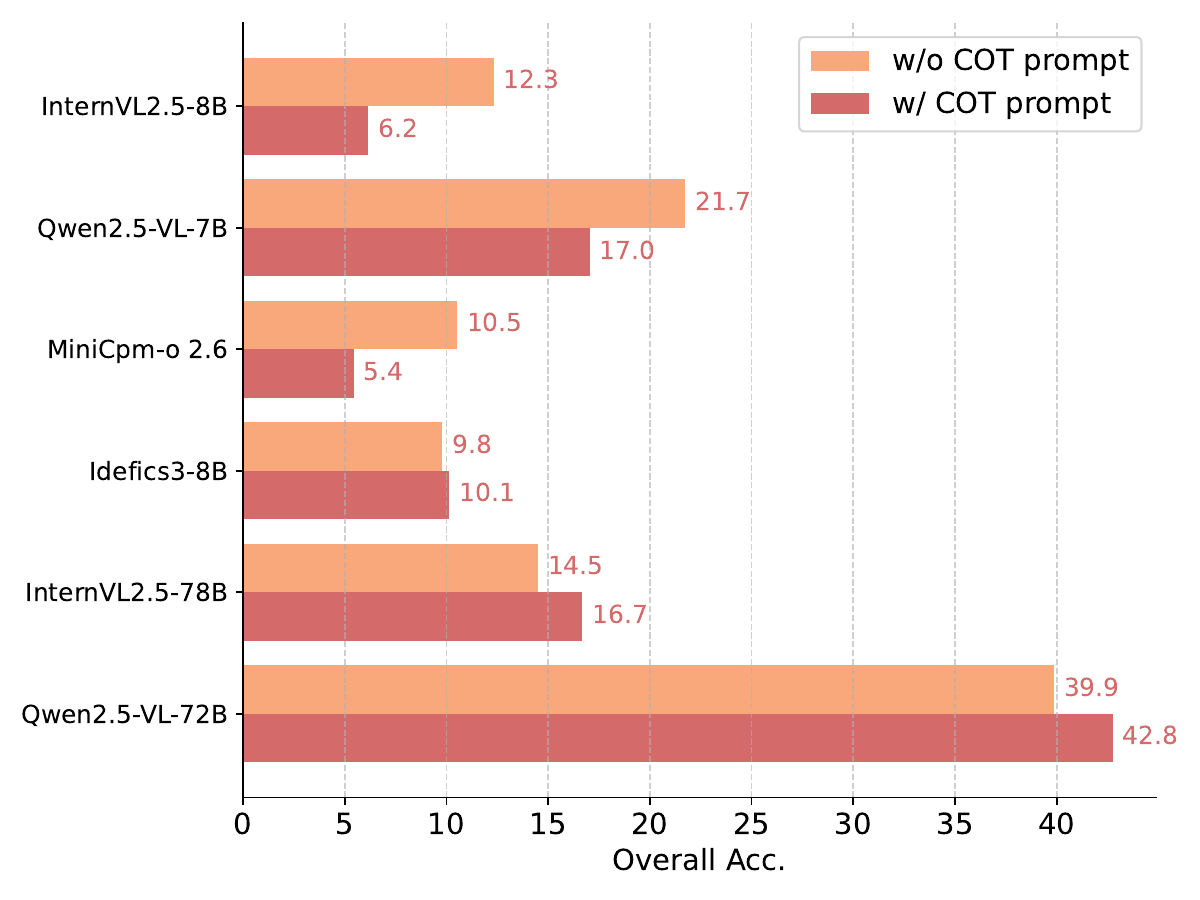}
\caption{Comparison of the overall accuracy of VLMs with and without CoT prompt.}
\label{COT}
\end{figure}

In the case of single-image input, increasing the image resolution typically enhances model performance~\cite{llavaonevision, Internvl2.0, monkey},. However, with multi-page image input, the large number of visual tokens present a significant challenge for the model in extracting useful information. Further increasing the image resolution also increases the number of visual tokens. How do the rich information and large number of visual tokens brought by high-resolution images affect the model's performance? To explore this question, we conducted experiments using four VLMs. The DPI of PDF-rendered JPEG images was gradually increased from 36 to 180. As shown in Figure \ref{resolution}, all four models exhibited steady performance improvements as the DPI increased to 108. However, when DPI exceeds 108, the model's performance fluctuates and even declines as DPI increases. This suggests that while moderately increasing resolution can benefit model performance for multi-image inputs, the model struggles to handle images with excessively high resolution.

\subsection{Does CoT Help in Answering Cross-Source Reasoning Questions?}


Figure \ref{COT} presents the impact of CoT prompt on our benchmark. Specifically, we designed CoT prompt, as outlined in the supplementary material, and evaluated the performance of the VLMs with and without CoT prompt. Our experiments revealed that smaller models experienced a notable performance decline when using CoT prompt. To explore this further, we evaluated larger models, such as InternVL2.5-78B and Qwen2.5-VL-72B, and observed that their performance improved with the use of CoT prompt. We hypothesize that the smaller models may struggle with severe hallucinations when confronted with a high number of visual tokens, or they may fail to identify sufficient sources of evidence in tasks requiring cross-source reasoning, which leads to a significant drop in CoT performance. In contrast, the larger models appear better equipped to handle the complexity of visual tokens and more effectively retrieve relevant evidence for coherent chain-of-thought reasoning. Interestingly, for the Idefics3-8B model, CoT prompt led to a slight improvement in performance. A detailed review of the Idefics3-8B reasoning process revealed that it still delivered answers directly without engaging in reasoning under the CoT prompt. These results highlight the varying effects of CoT across models of different sizes, particularly in their capacity to leverage cross-source in multi-page scientific papers. They suggest that the ability to effectively extract and integrate cross-source information from numerous visual tokens is key to improving performance with CoT prompt.

\subsection{Error Analysis}

\begin{figure}
\centering
\includegraphics[width=0.5\textwidth]{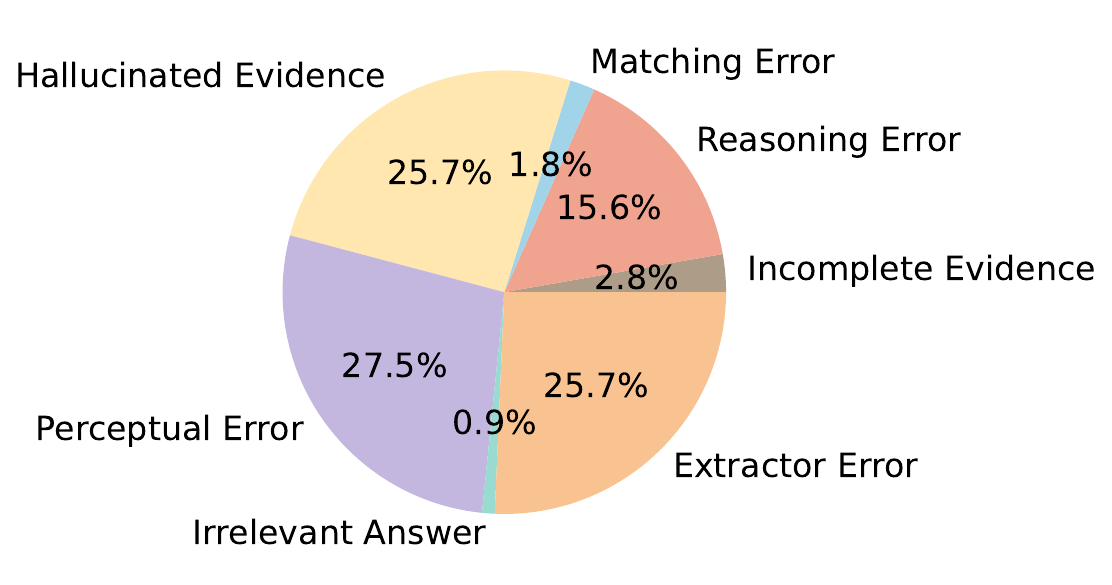}
\caption{Error distribution over 109 annotated GPT-4o errors.}
\label{error_analogy}
\end{figure}

\begin{figure}
\centering
\includegraphics[width=0.45\textwidth]{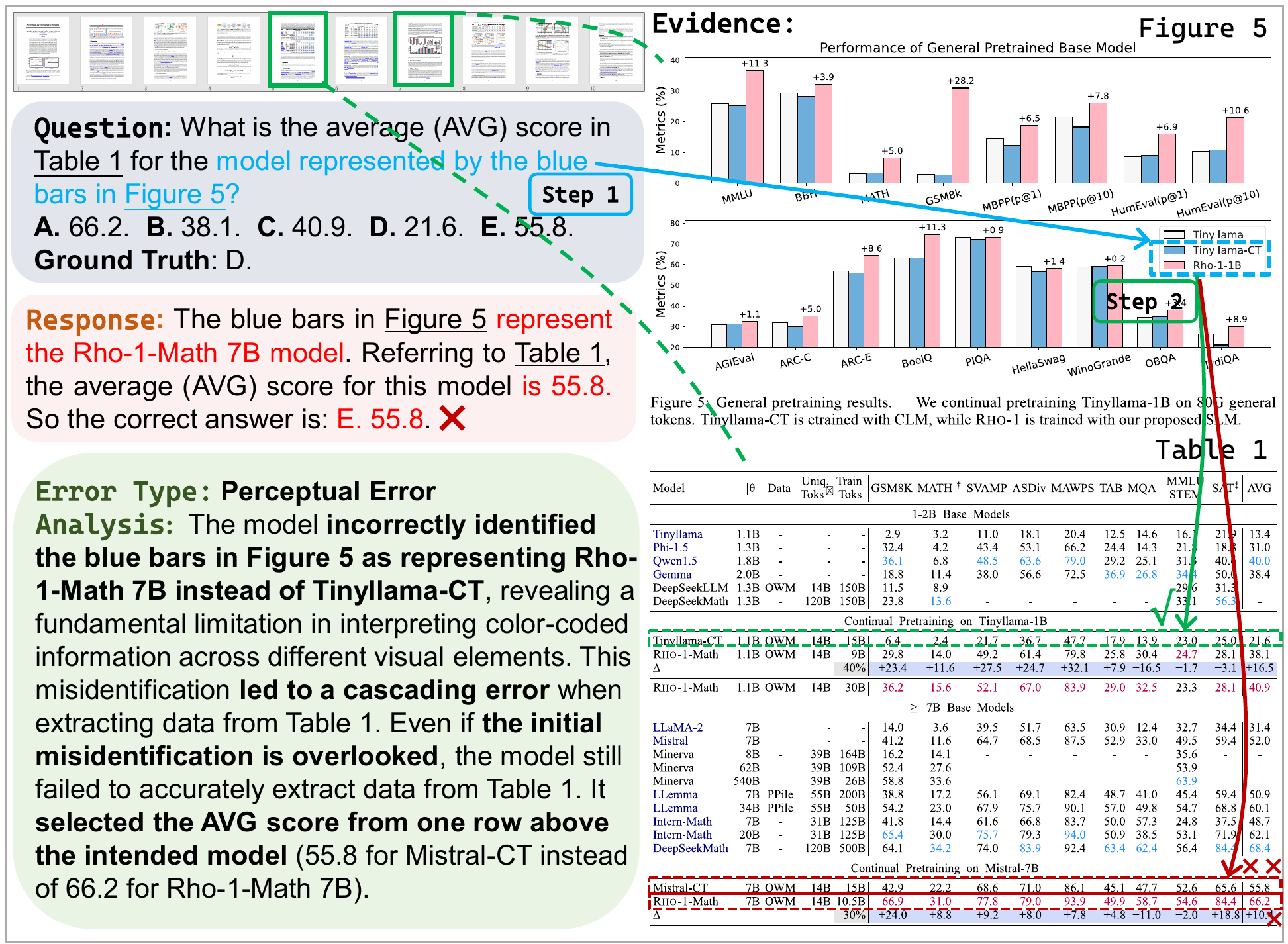}
\caption{An example of a perceptual error is observed in GPT-4o, where the model fails to accurately interpret both the figure and the table, resulting in unsuccessful cross-source reasoning. The correct reasoning process is outlined in Step 1 and Step 2.}
\label{error_example}
\end{figure}

We further conducted an error analysis to identify the bottlenecks in current multimodal models when handling cross-source reasoning tasks. Specifically, we selected 109 instances where GPT-4o provided incorrect answers and performed a manual examination. The errors were categorized into the following types: 1) \textbf{Hallucinated Evidence:} The model failed to locate the necessary evidence to answer the question. 2) \textbf{Incomplete Evidence:} The evidence retrieved was insufficient. 3) \textbf{Perceptual Error:} The model misinterpreted graphs, symbols, colors, etc. 4) \textbf{Extractor Error:} The model extracted values that did not exist in the table or extracted values from misaligned tables. 5) \textbf{Reasoning Error:} The model made incorrect inferences based on the retrieved information. 6) \textbf{Irrelevant Answer:} The model’s answer was unrelated to the question. 7) \textbf{Matching Error.} Rule-based methods failed to extract the correct response from the model. The distribution of these errors is shown in Figure \ref{error_analogy}. Our findings reveal that the most prevalent error is perceptual error, followed by extractor error, highlighting the model's difficulty in accurately retrieving relevant information from multi-page images. Hallucinated and incomplete evidence together account for 28.5\% of the errors, further underscoring the substantial room for improvement in current VLMs in leveraging cross-source information for reasoning. Figure \ref{error_example} presents our analysis of a perceptual error, offering insights into its underlying causes. 

\section{Conclusion}

In this work, we present MMCR, a benchmark for evaluating VLMs' ability to perform cross-source reasoning in scientific papers. Our extensive experiments reveal that current VLMs struggle to identify complete and accurate evidence required for reasoning from cross-source information in scientific papers. We hope that the development of this benchmark will promote advancement in VLMs' capabilities to perform reasoning utilizing cross-source information.

\noindent\textbf{Acknowledgement.} This work was funded by National Natural Science Foundation of China under No. 62306046. We thank MolarData for their technical and resource support.


{
    \small
    \bibliographystyle{ieeenat_fullname}
    \bibliography{main}
}

\clearpage

\appendix
\section*{Appendix}
\begin{appendices}
    \renewcommand{\thefigure}{S.\arabic{figure}}
    \renewcommand{\thetable}{S.\arabic{table}}
    

\section{Benchmark Details}
We appreciate the reviewer's thoughtful comment on MMCR's classification as a reasoning benchmark. We respectfully maintain that MMCR  tests reasoning capabilities as it aligns with the reviewer's cited definition of reasoning as "multi-step/multi-hop question answering."Our definition of "cross-source reasoning" in MMCR refers to questions that require synthesizing information from multiple sources within scientific papers to derive answers that cannot be obtained from any single source alone. 

Taking the question in Figure S.7 as an example, the reasoning process in MMCR directly parallels HotpotQA's definition of "inferring the bridge entity to complete the 2nd-hop question." As illustrated in Figure S.7, answering MMCR questions typically requires first inferring which specific figure or table is being referenced through textual descriptions (e.g., "the t-SNE visualization of CLIP encoding features"). This constitutes the first reasoning hop and establishes the critical bridge entity. Only after successfully identifying this bridge entity can the model proceed to the second hop, extracting relevant information from the identified sources and synthesizing it to derive the answer.

\begin{figure}[htbp]
    \centering
    \subfigure[]{\includegraphics[width=\linewidth]{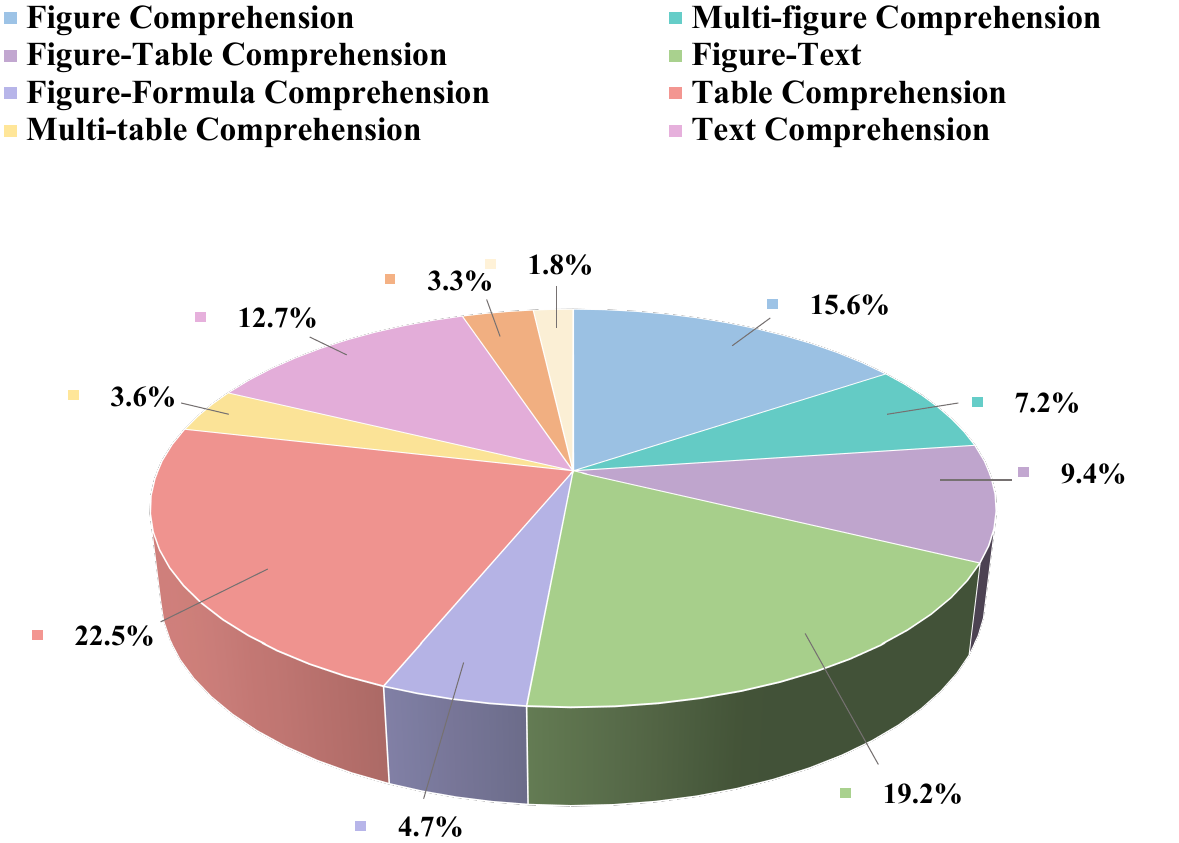}}
    \subfigure[]{\includegraphics[width=\linewidth]{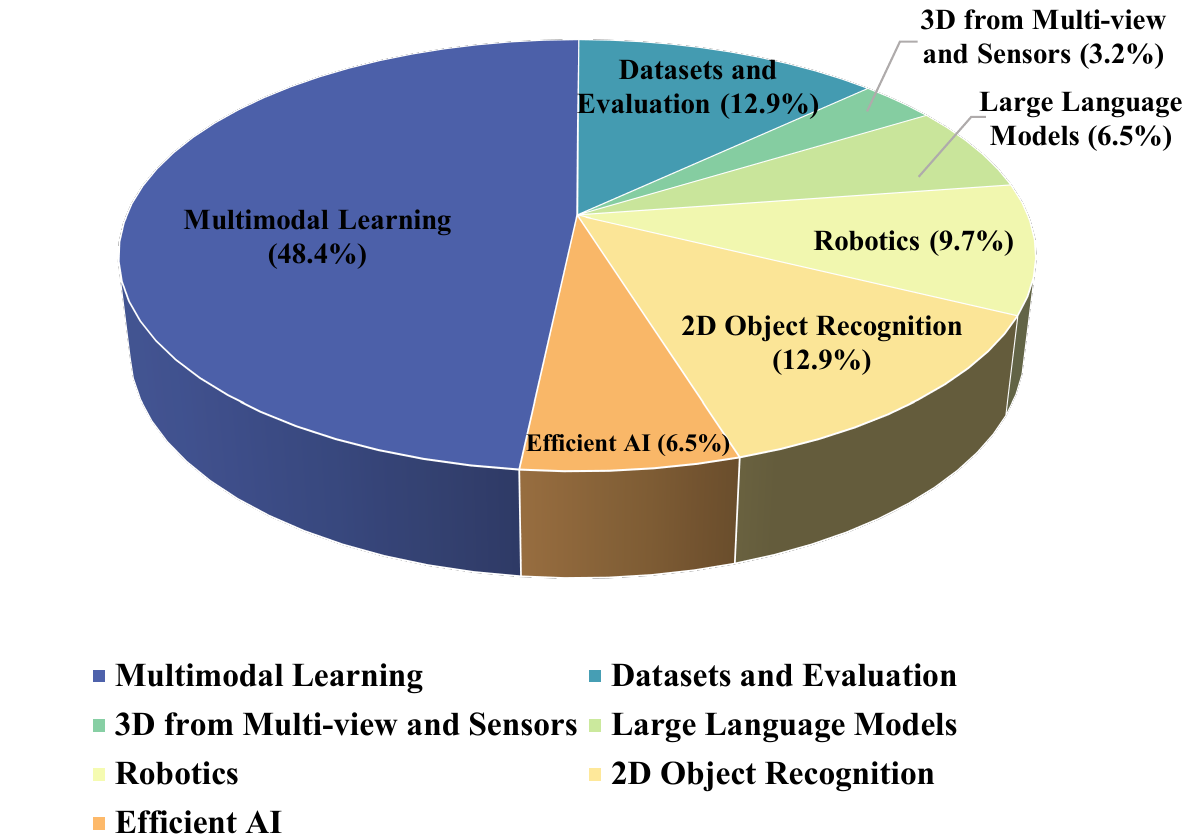}}
    \caption{Distribution of Questions by Evidence Types and Research Domains. (a) Percentage distribution across ten evidence source types. (b) Distribution across seven AI research subjects.}
    \label{fig:benchmark_pie_chart}
\end{figure}

\begin{figure*}
    \centering
    \includegraphics[width=\linewidth]{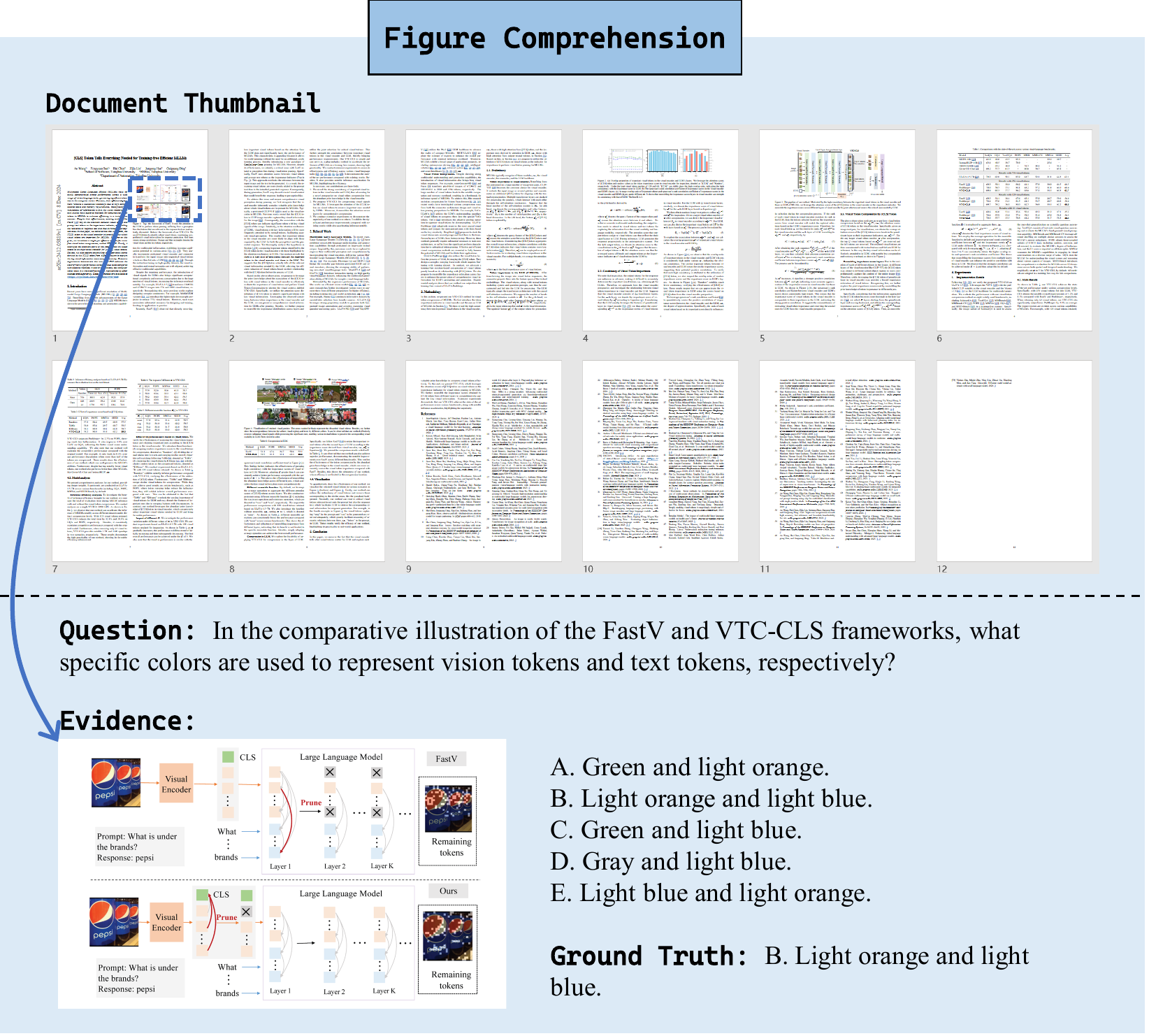}
    \caption{\textbf{The demo of figure comprehension}.}
    \label{fig:figure-comprehension}
\end{figure*}

\begin{figure*}
    \centering
    \includegraphics[width=\linewidth]{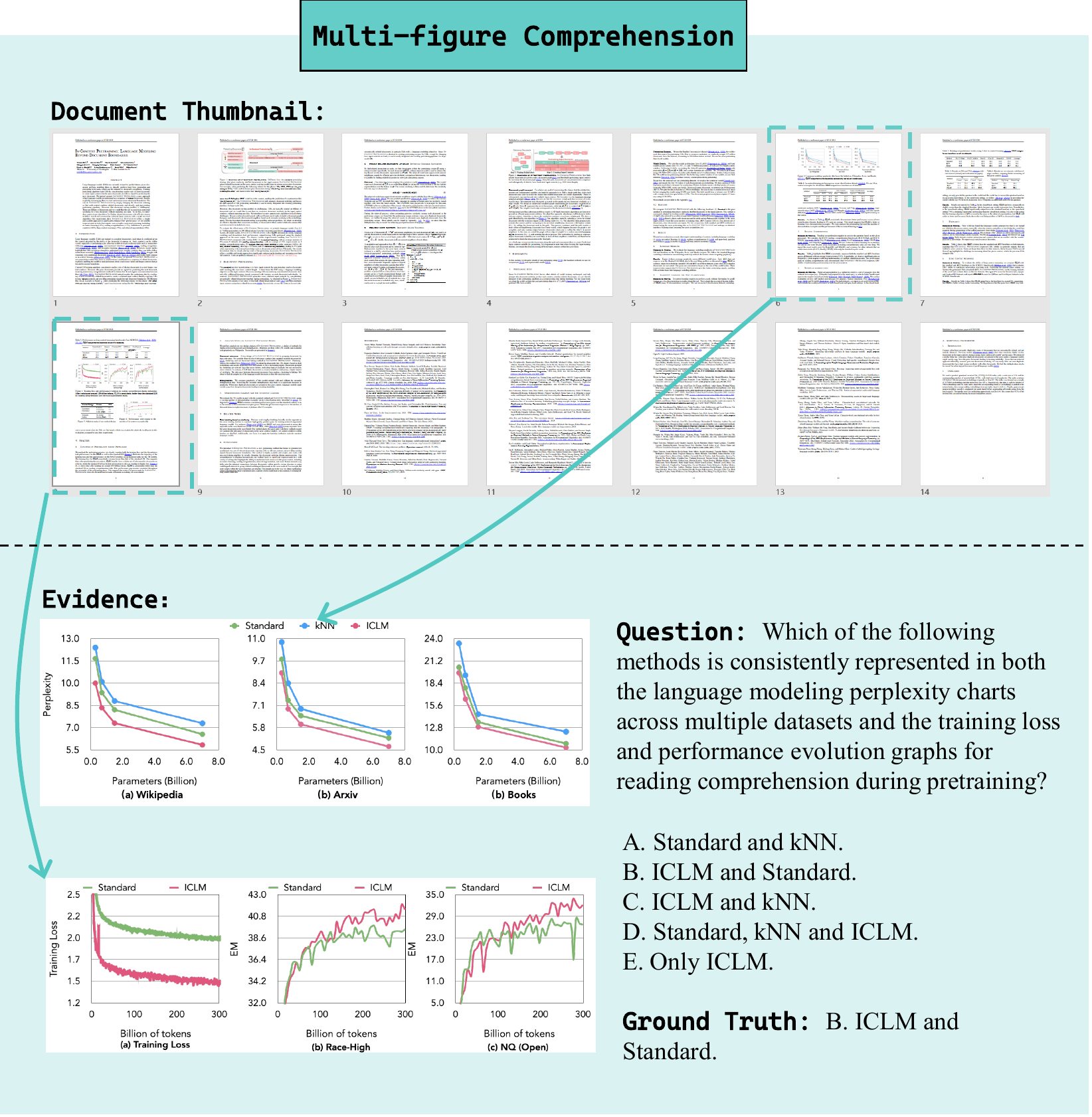}
    \caption{\textbf{The demo of multi-figure comprehension}.}
    \label{fig:multi-figure-comprehension}
\end{figure*}

\begin{figure*}
    \centering
    \includegraphics[width=\linewidth]{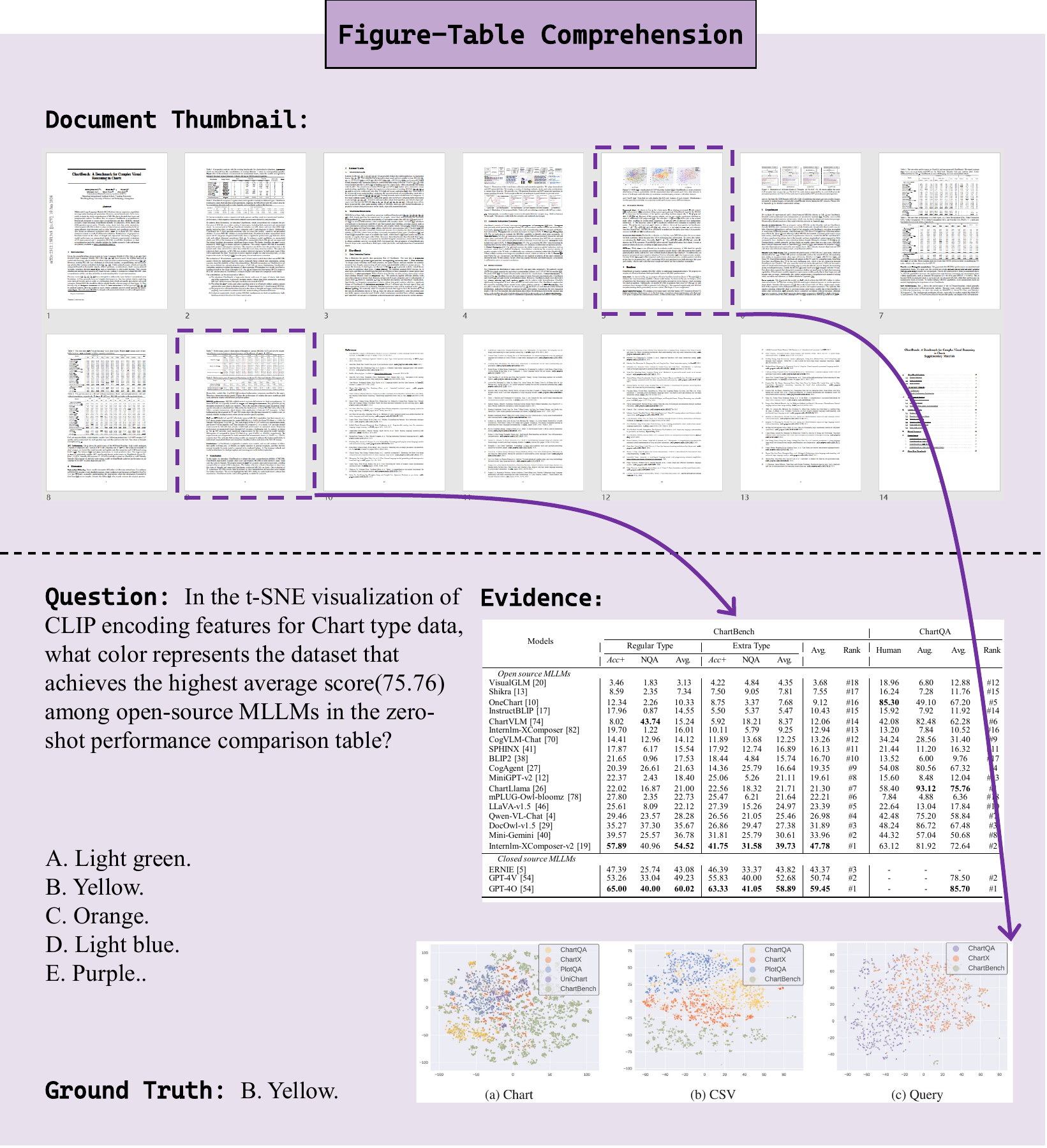}
    \caption{\textbf{The demo of figure-table comprehension}.}
    \label{fig:figure-table-comprehension}
\end{figure*}

\begin{figure*}
    \centering
    \includegraphics[width=\linewidth]{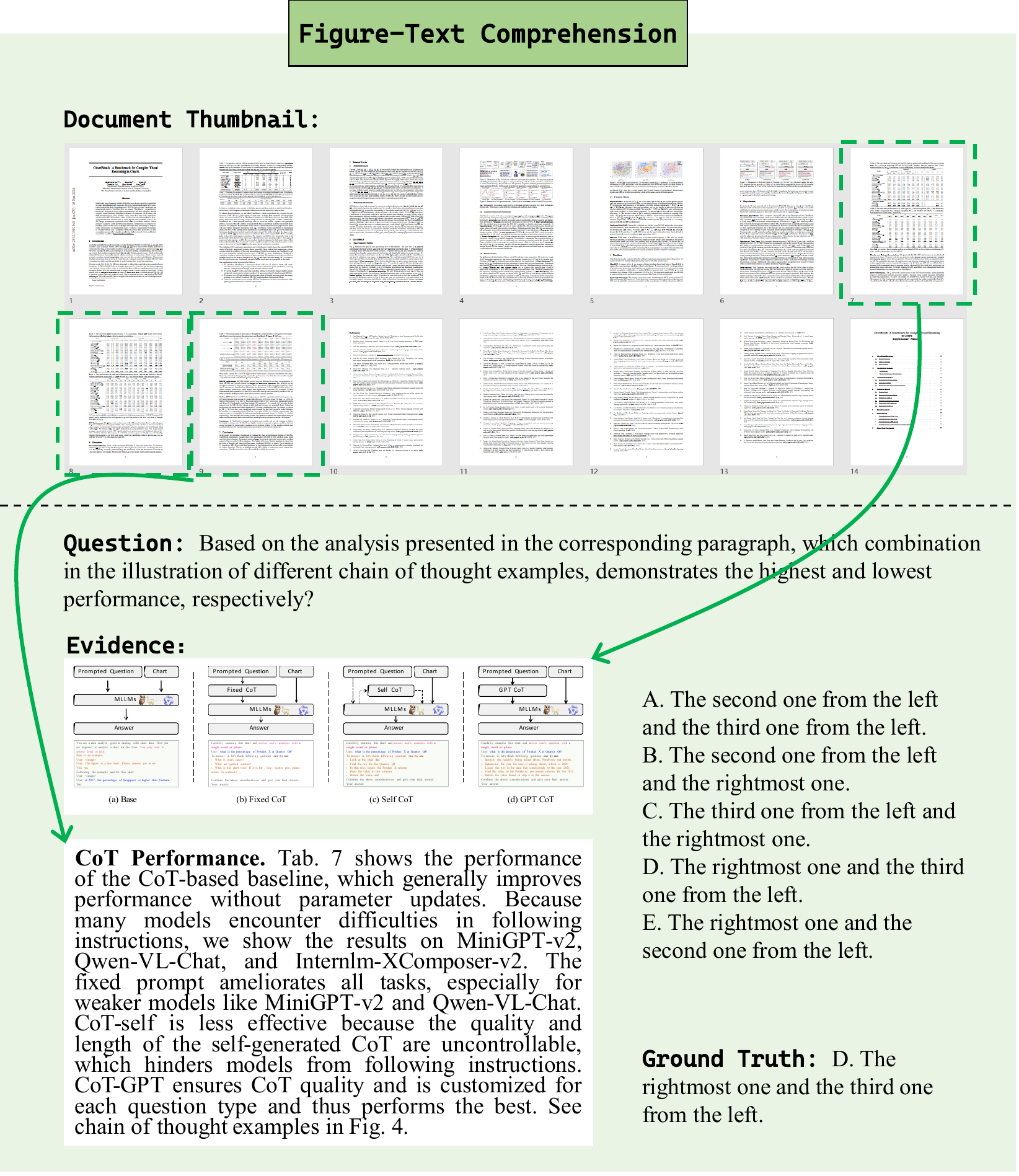}
    \caption{\textbf{The demo of figure-text comprehension}.}
    \label{fig:figure-text-comprehension}
\end{figure*}

\begin{figure*}
    \centering
    \includegraphics[width=\linewidth]{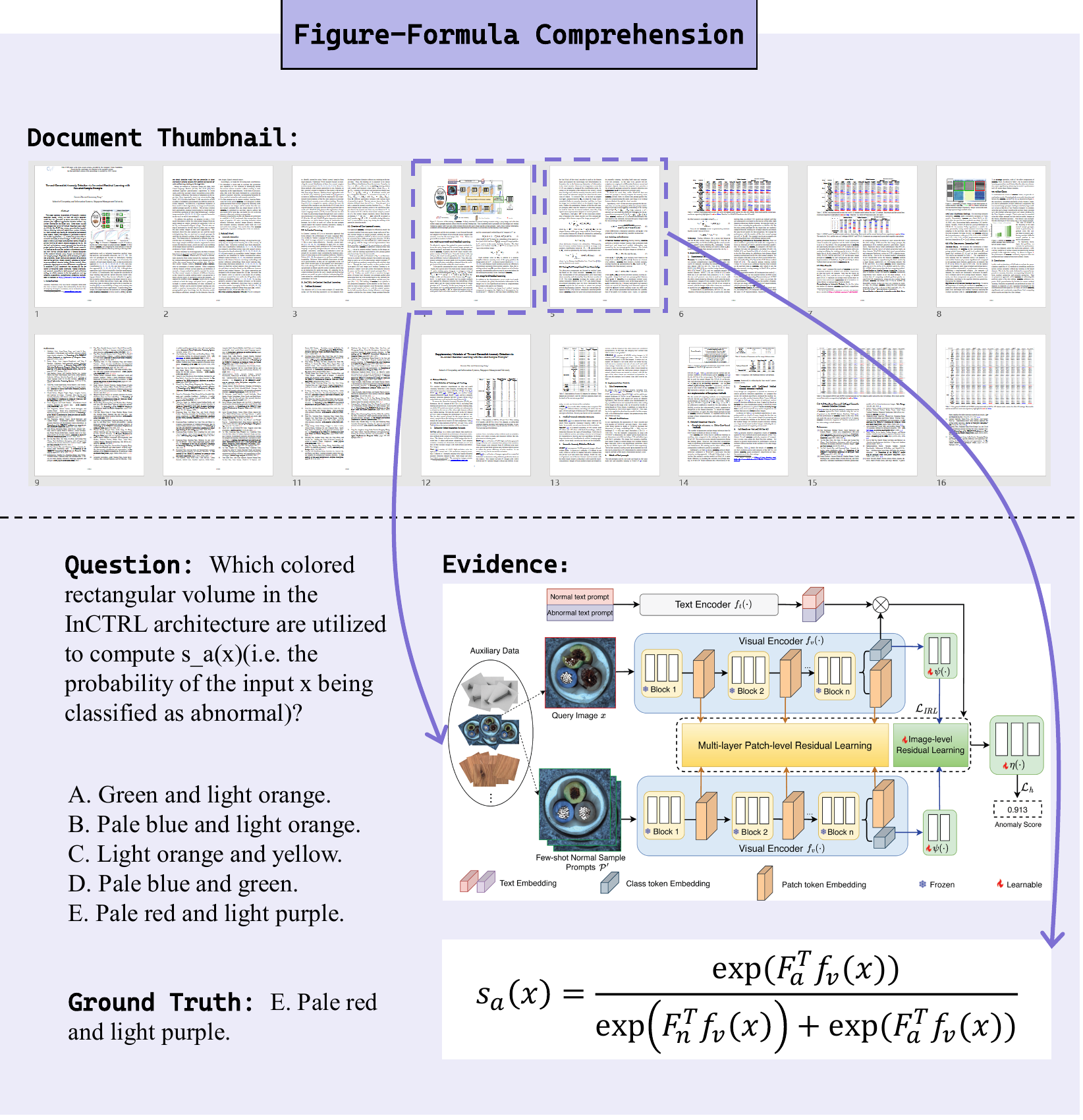}
    \caption{\textbf{The demo of figure-formula comprehension}.}
    \label{fig:figure-formula-comprehension}
\end{figure*}

\begin{figure*}
    \centering
    \includegraphics[width=\linewidth]{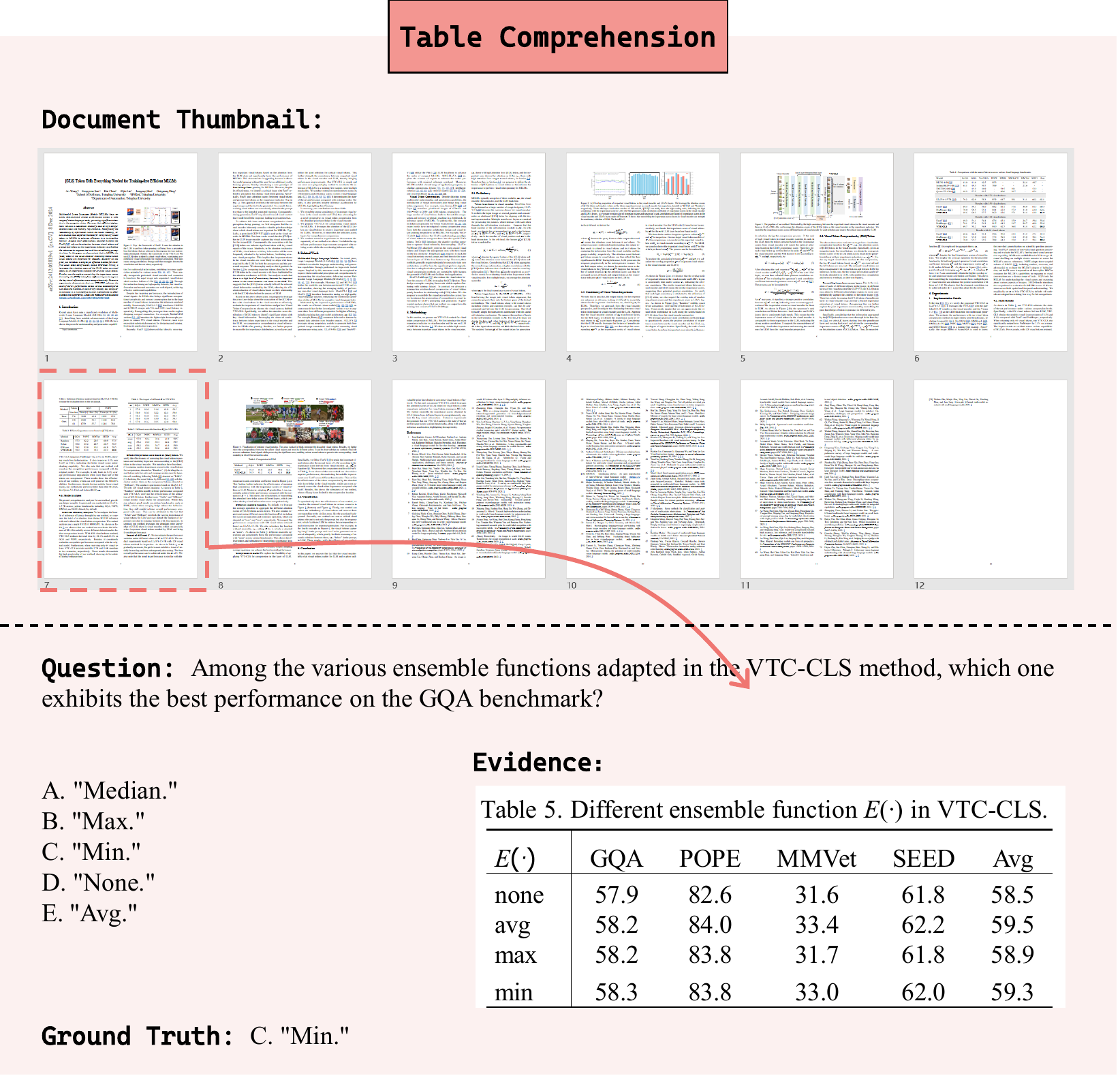}
    \caption{\textbf{The demo of table comprehension}.}
    \label{fig:table-comprehension}
\end{figure*}

\begin{figure*}
    \centering
    \includegraphics[width=0.98\linewidth]{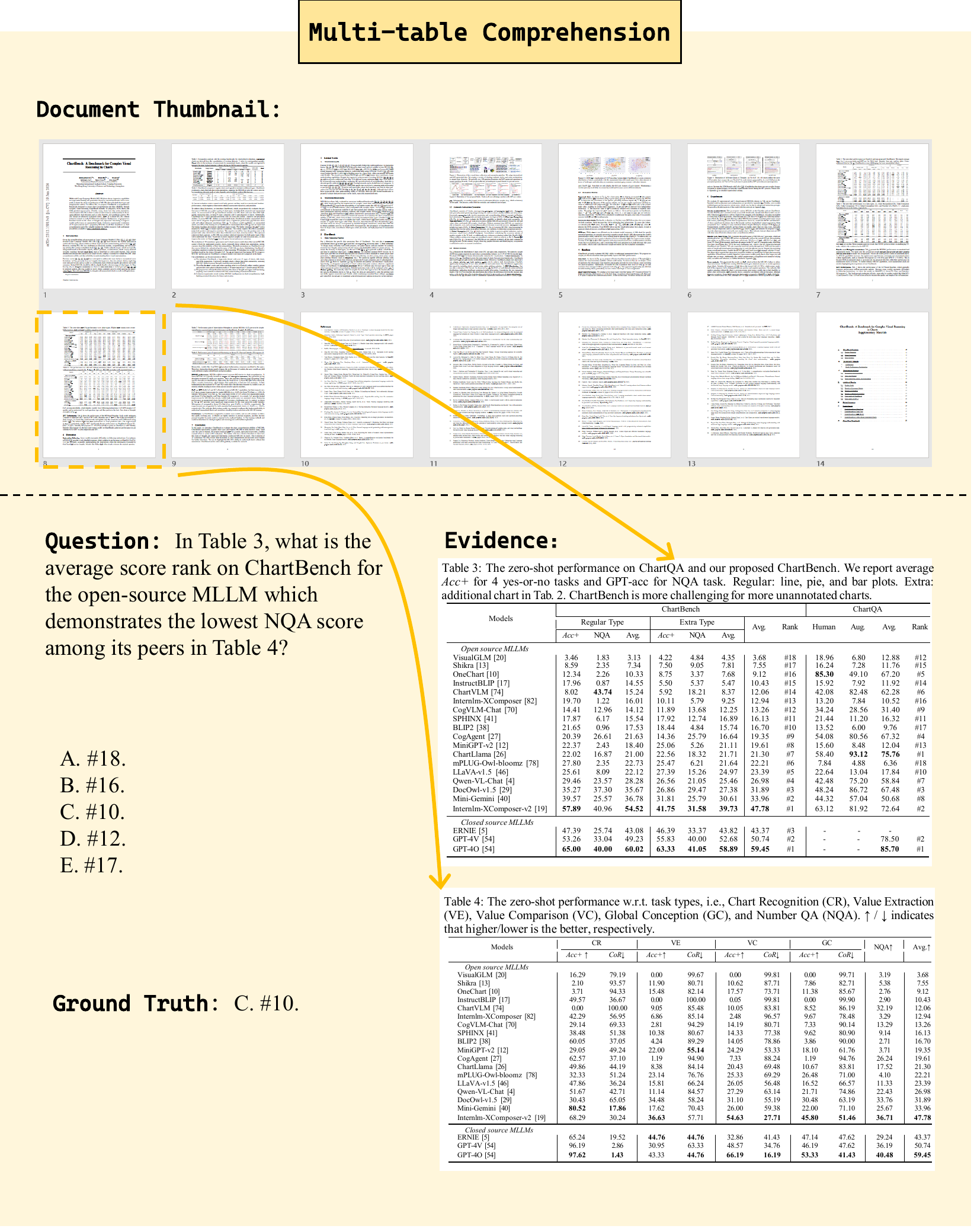}
    \caption{\textbf{The demo of multi-table comprehension}.}
    \label{fig:multi-table-comprehension}
\end{figure*}

\begin{figure*}
    \centering
    \includegraphics[width=\linewidth]{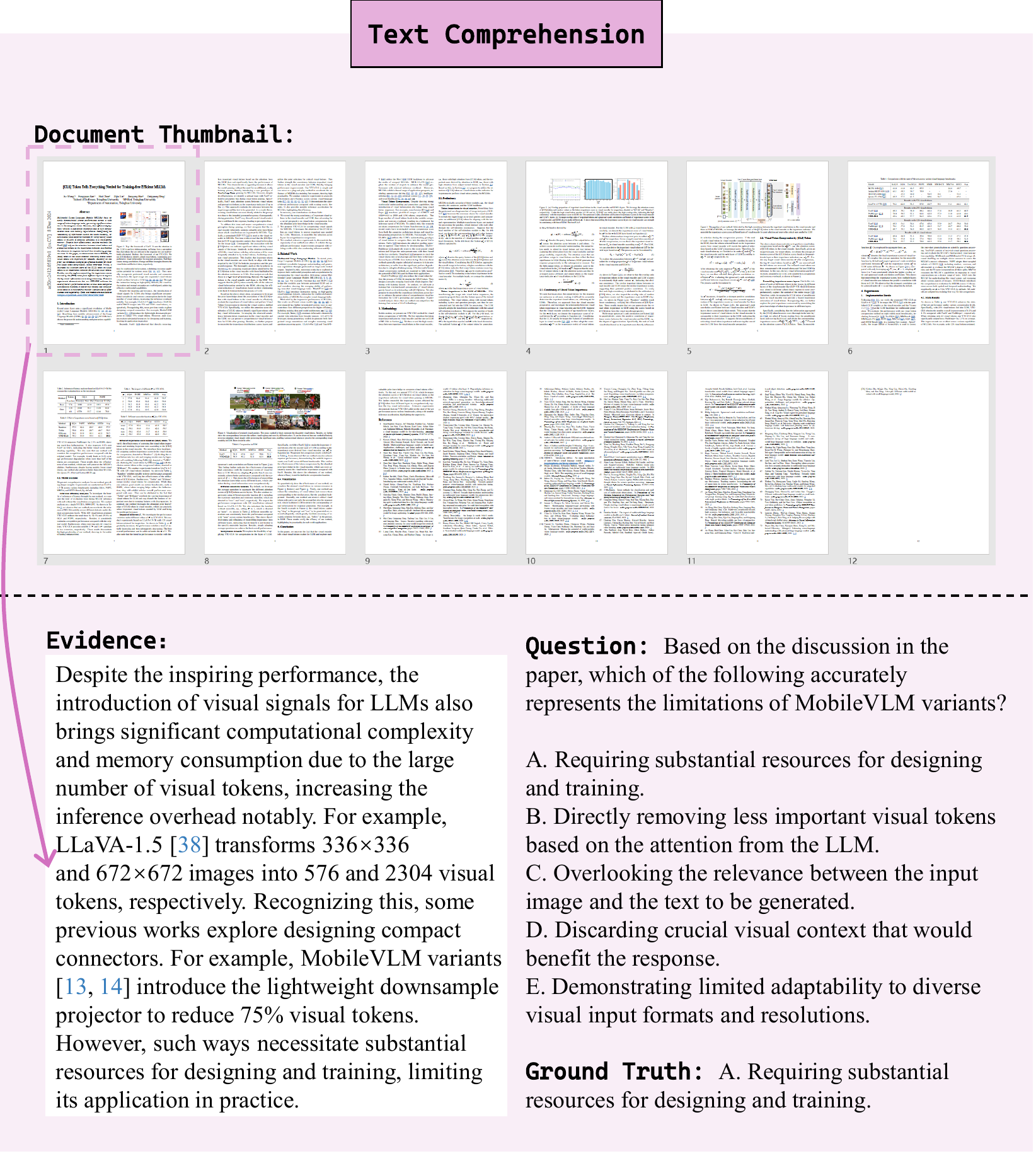}
    \caption{\textbf{The demo of text comprehension}.}
    \label{fig:text-comprehension}
\end{figure*}

\begin{figure*}
    \centering
    \includegraphics[width=\linewidth]{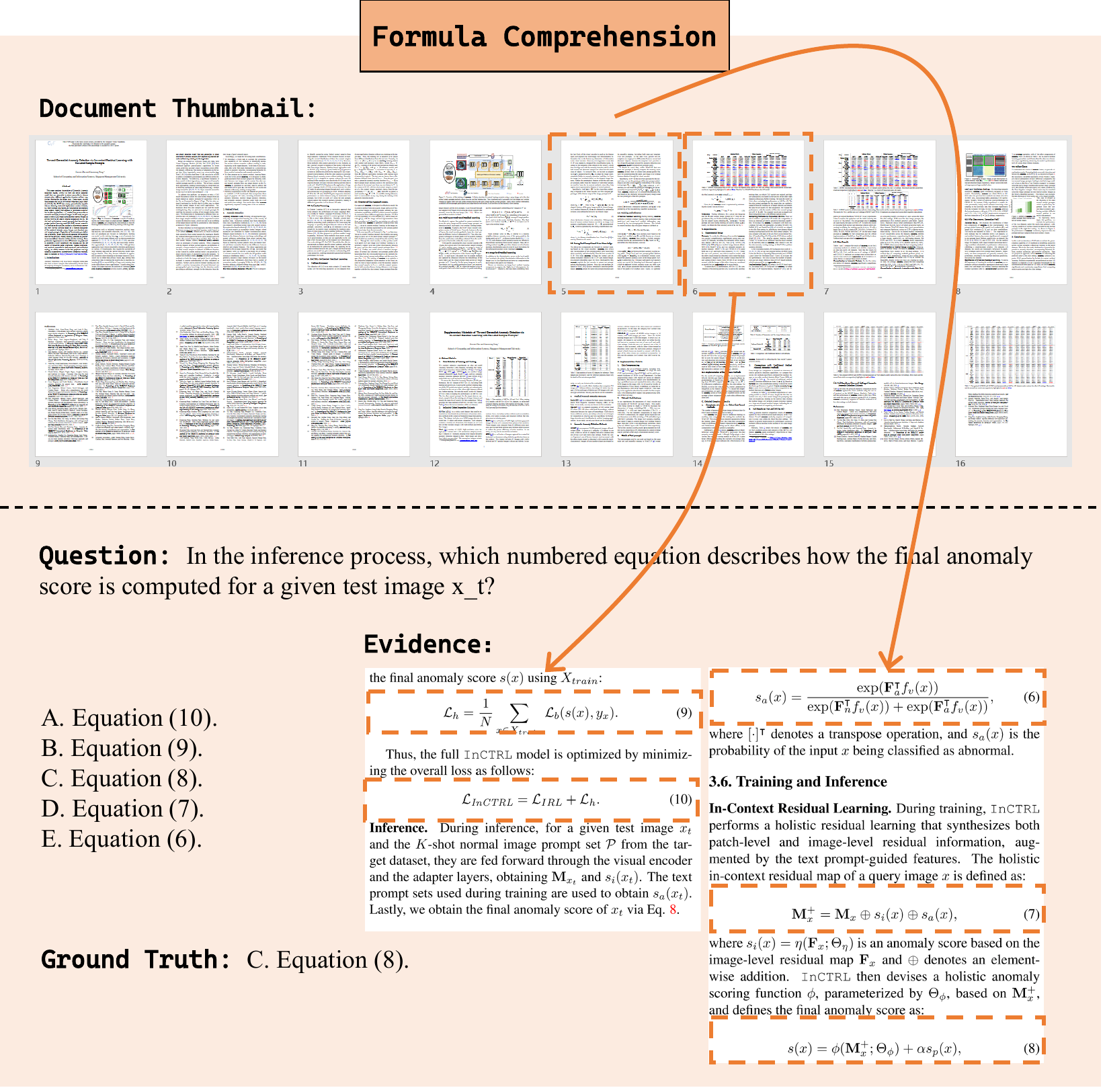}
    \caption{\textbf{The demo of formula comprehension}.}
    \label{fig:formula-comprehension}
\end{figure*}

\begin{figure*}
    \centering
    \includegraphics[width=\linewidth]{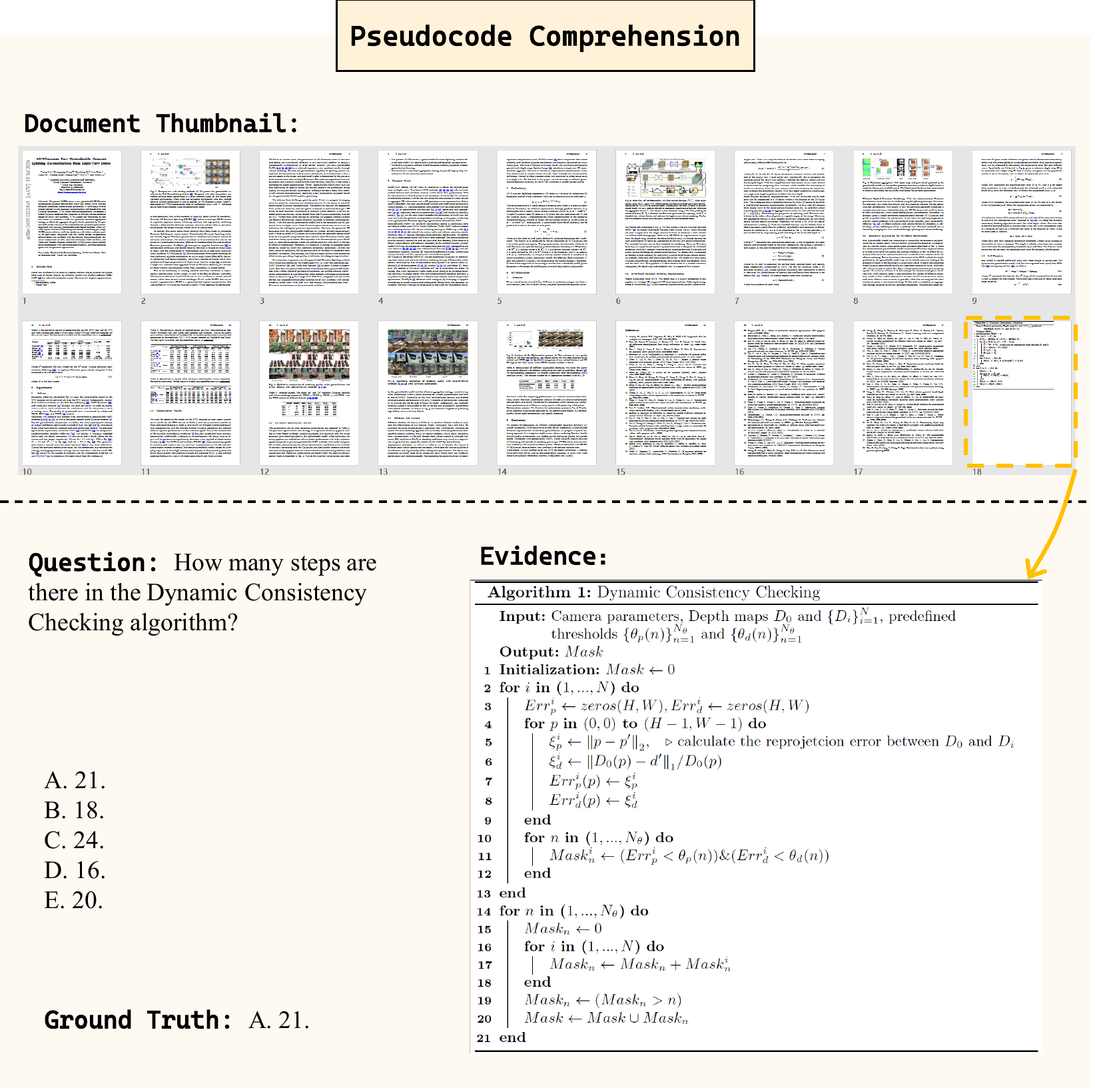}
    \caption{\textbf{The demo of pseudocode comprehension}.}
    \label{fig:pseudocode-comprehension}
\end{figure*}

The reasoning complexity is further amplified by scientific papers' high information density (19 pages average), input as pure images rather than OCR-processed text, and questions requiring numerical reasoning and calculations.
The benchmark comprises scientific papers across seven academic subjects of artificial intelligence, with questions systematically categorized into ten distinct types based on their required evidence sources. The distribution of these categories is illustrated in \cref{fig:benchmark_pie_chart}. Representative examples demonstrating each evidence type are presented in \cref{fig:figure-comprehension,fig:multi-figure-comprehension,fig:figure-table-comprehension,fig:figure-text-comprehension,fig:figure-formula-comprehension,fig:table-comprehension,fig:multi-table-comprehension,fig:text-comprehension,fig:formula-comprehension,fig:pseudocode-comprehension}.

\section{Evaluation Details}
\subsection{Evaluation Prompt}
\Cref{fig:evaluation-prompts} presents the prompts with and without the use of Chain-of-Thought (CoT). For InternVL2.5, we employed the official CoT prompt released by the developers. For the remaining benchmark models—MiniCpm-o 2.6, Qwen2.5-VL, and Idefics3—we implemented a unified CoT prompt to ensure methodological consistency across experiments.
\begin{figure}[htbp]
\centering
\definecolor{skyblue}{RGB}{91,155,213}
\definecolor{grassgreen}{RGB}{112,173,71}
\begin{tcolorbox}[colback=lightgray!20,colframe=skyblue, title=Evaluation Prompts: CoT Prompt]
\textbf{CoT Prompt for InternVL-2.5:} \\
``Your task is to answer the question below. Give step by step reasoning before you answer, and when you're ready to answer, please use the format: \\
`\textbackslash Final answer: ... \textbackslash'\\ 
Question: \{question\}"

\vspace{0.2cm}
\textbf{CoT Prompt for MiniCpm-o 2.6, Qwen2.5-VL, Idefics3:} \\
``Carefully read the following multichoice question, solve it step by step and finally pick the option associated with the correct answer in the format of `Answer: Selected option."

\vspace{0.2cm}

\textbf{w/o CoT Prompt:}\\
``Please select the correct answer from the options above."
\end{tcolorbox}

\caption{Evaluation Prompt}
\label{fig:evaluation-prompts}
\end{figure}

\subsection{Answer Option Inference Details for LLM Responses}
\subsubsection{Implementation Details}
Unlike existing benchmarks that employ LLM-based methods for open-ended response extraction, our benchmark utilizes a heuristic rule-based approach for multiple-choice answer inference.

The rule-based approach for multiple-choice answer inference comprises two stages: primary option-based inference, followed by text-based inference as a fallback strategy. Specifically, the option-based inference method counts the occurrence of option identifiers (A, B, C, etc.) in the response. A valid inference is made when exactly one option identifier is detected. When option-based inference fails, the text-based inference serves as a fallback mechanism. It converts both the model response and choice contents to lowercase before searching for exact matches of choice content within the response. This method succeeds only when precisely one choice content is found in the processed response.

This two-stage approach ensures robust answer extraction while maintaining high precision through strict matching criteria. When both methods fail to identify a unique answer, false will be returned to indicate inference failure.

\subsubsection{Existing Problems}
The rule-based approach for multiple-choice answer inference offers efficiency by eliminating additional LLM calls. However, it occasionally fails to accurately extract responses despite correct model reasoning. We categorize such cases as \textit{Matching Errors}. As shown in \cref{tab:distribution of matching errors}. The error distribution analysis demonstrates that extraction failures represent a negligible proportion of the total errors, with only two instances identified across all cases. \Cref{fig:matching_error} illustrates one representative example of such a Matching Error.
\begin{table}[htbp]
    \centering
     \renewcommand{\arraystretch}{1.2}
    \begin{tabular}{c|c|c}
    \toprule[1.5pt]
      \textbf{Matching errors} & \textbf{Total error cases} & \textbf{Error Rate}\\
    \hline
         2&  109& 1.8\% \\
    \bottomrule[1.5pt]
    \end{tabular}
    \caption{ Example Distribution of Matching Errors in Response Extraction from GPT-4o.The table shows the proportion of matching errors among all error cases, demonstrating that extraction failures constitute only 1.8\% of total errors identified in our evaluation.}
    \label{tab:distribution of matching errors}
\end{table}

 \begin{figure*}
    \centering
    \includegraphics[width=\linewidth]{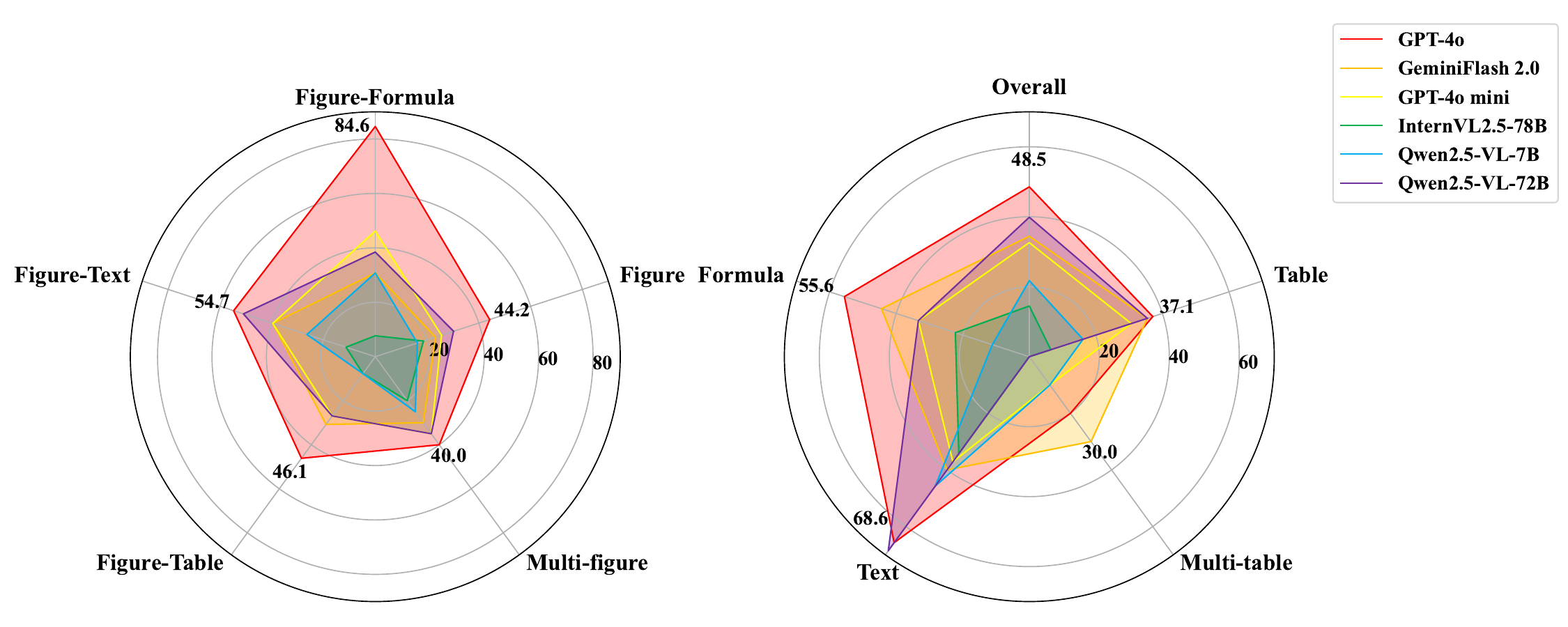}
    \caption{Fine-grained results on various evidence source types.}
    \label{fig:radar_chart_for_evidence_type}
\end{figure*}
\section{Extended Analysis}
\subsection{Analysis of Error Cases}
\begin{figure*}
    \centering
    \includegraphics[width=\linewidth]{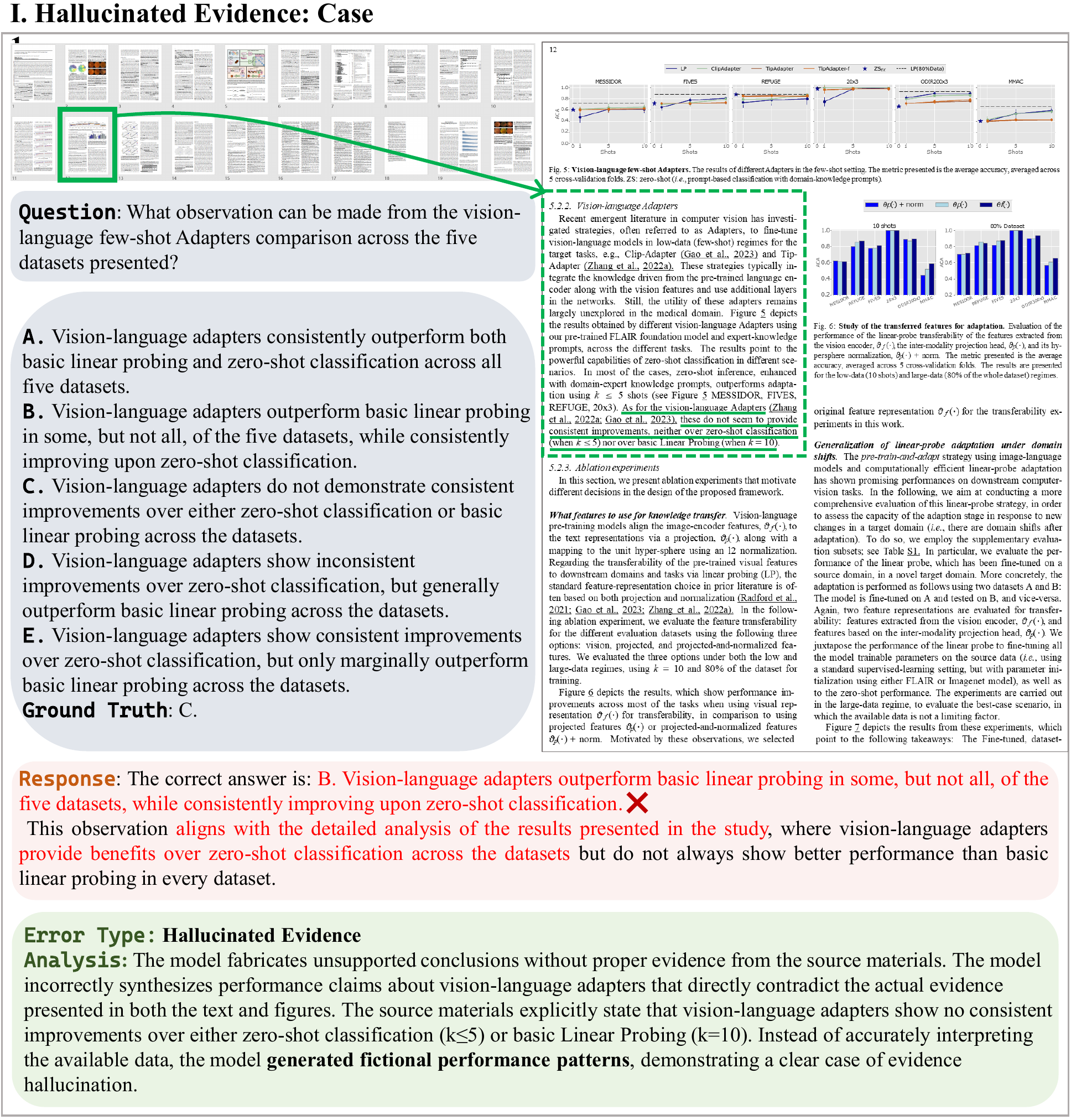}
    \caption{Illustration of a Hallucinated Evidence Error Case. The figure demonstrates how the model fabricates unsupported conclusions about vision-language adapter performance, contradicting the evidence highlighted in green from the source materials.}
    \label{fig:hallucinated_evidence}
\end{figure*}

\begin{figure*}
    \centering
    \includegraphics[width=\linewidth]{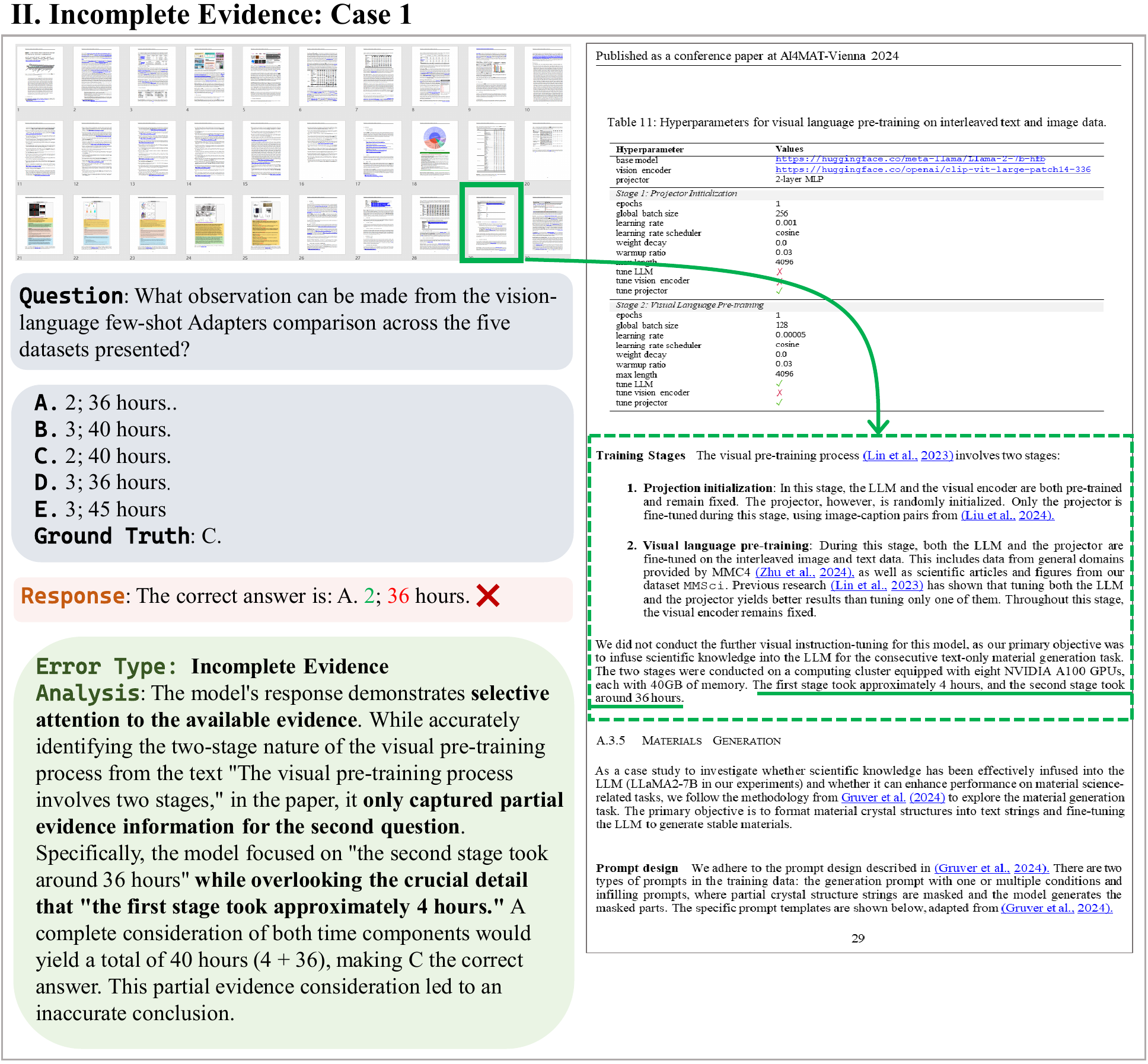}
    \caption{Illustration of an Incomplete Evidence Error Case. The model correctly identifies the two-stage nature of visual pre-training but overlooks crucial timing information, considering only the second stage duration (36 hours) while missing the first stage duration (4 hours).}
    \label{fig:incomplete_evidence}
\end{figure*}

\begin{figure*}
    \centering
    \includegraphics[width=\linewidth]{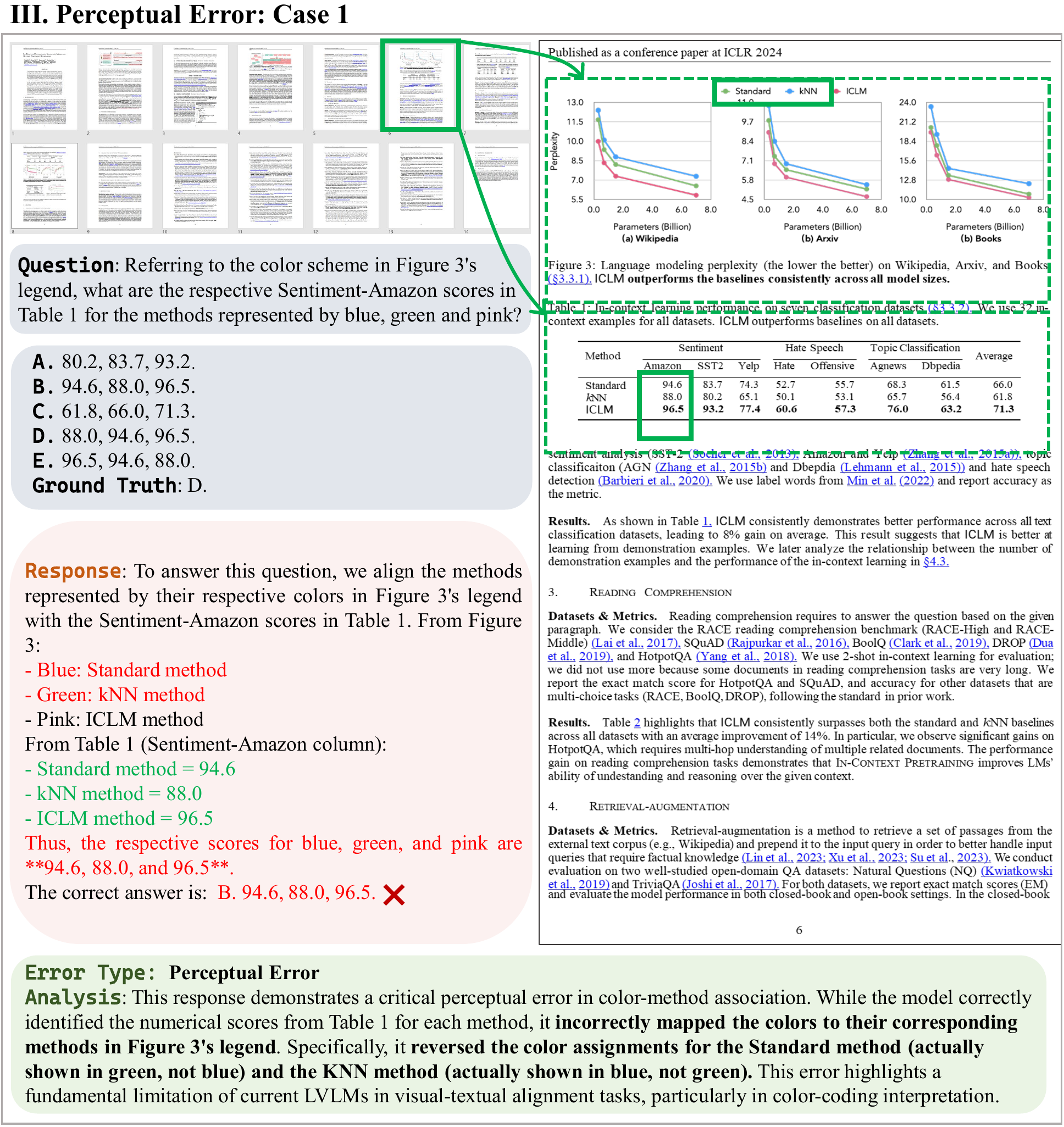}
    \caption{Illustration of a Perceptual Error Case. The figure demonstrates how the model correctly identifies numerical scores but incorrectly maps color codes to methods, mismatching the blue and green representations in the legend despite clear visual evidence.}
    \label{fig:perceptual_error}
\end{figure*}

\begin{figure*}
    \centering
    \includegraphics[width=\linewidth]{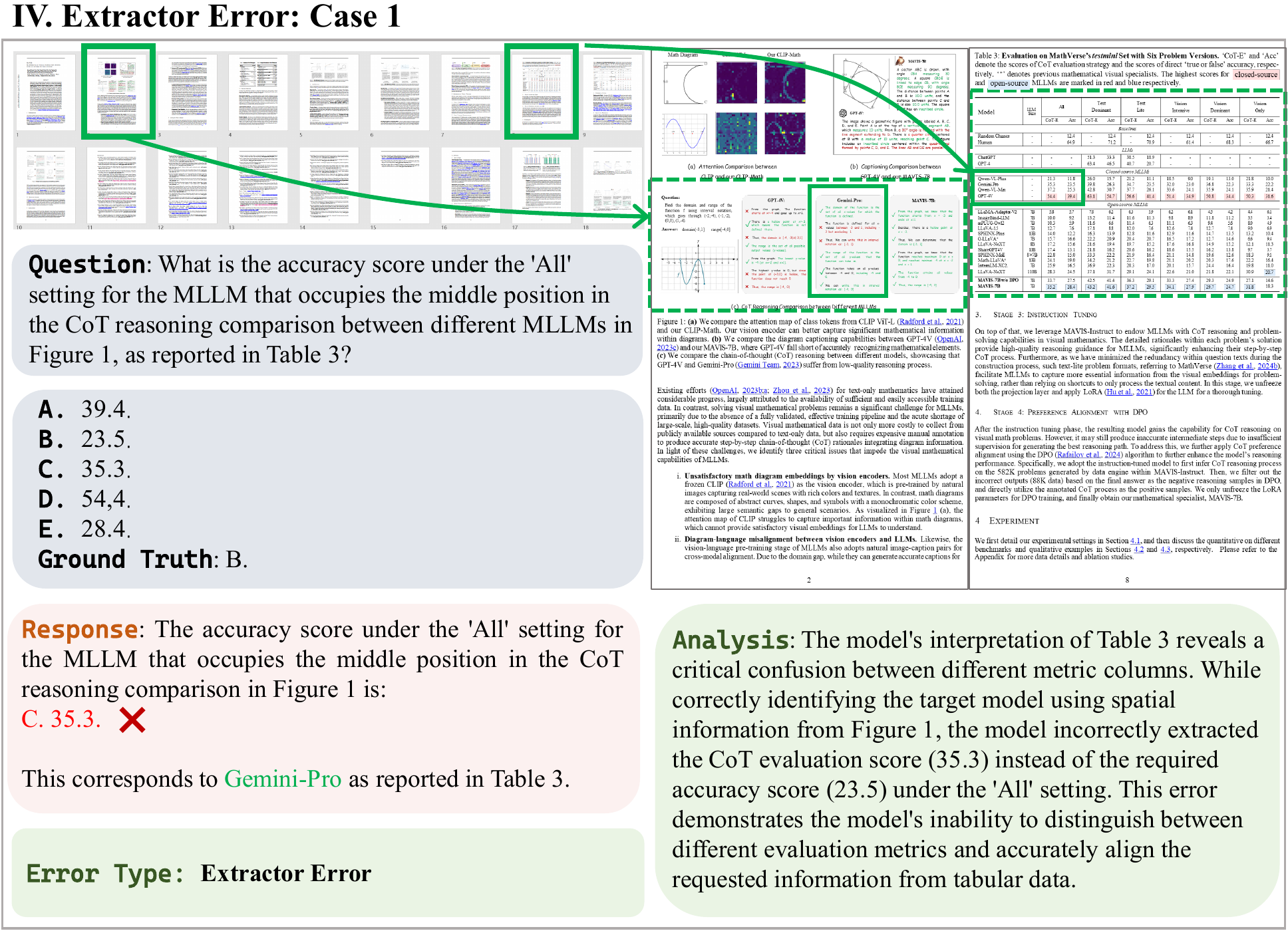}
    \caption{Illustration of an Extractor Error Case. The figure demonstrates how the model confuses different metric columns in Table 3, extracting the CoT evaluation score (35.3) instead of the correct accuracy score (23.5) despite accurately identifying the target model from spatial information.}
    \label{fig:extractor_error}
\end{figure*}

\begin{figure*}
    \centering
    \includegraphics[width=\linewidth]{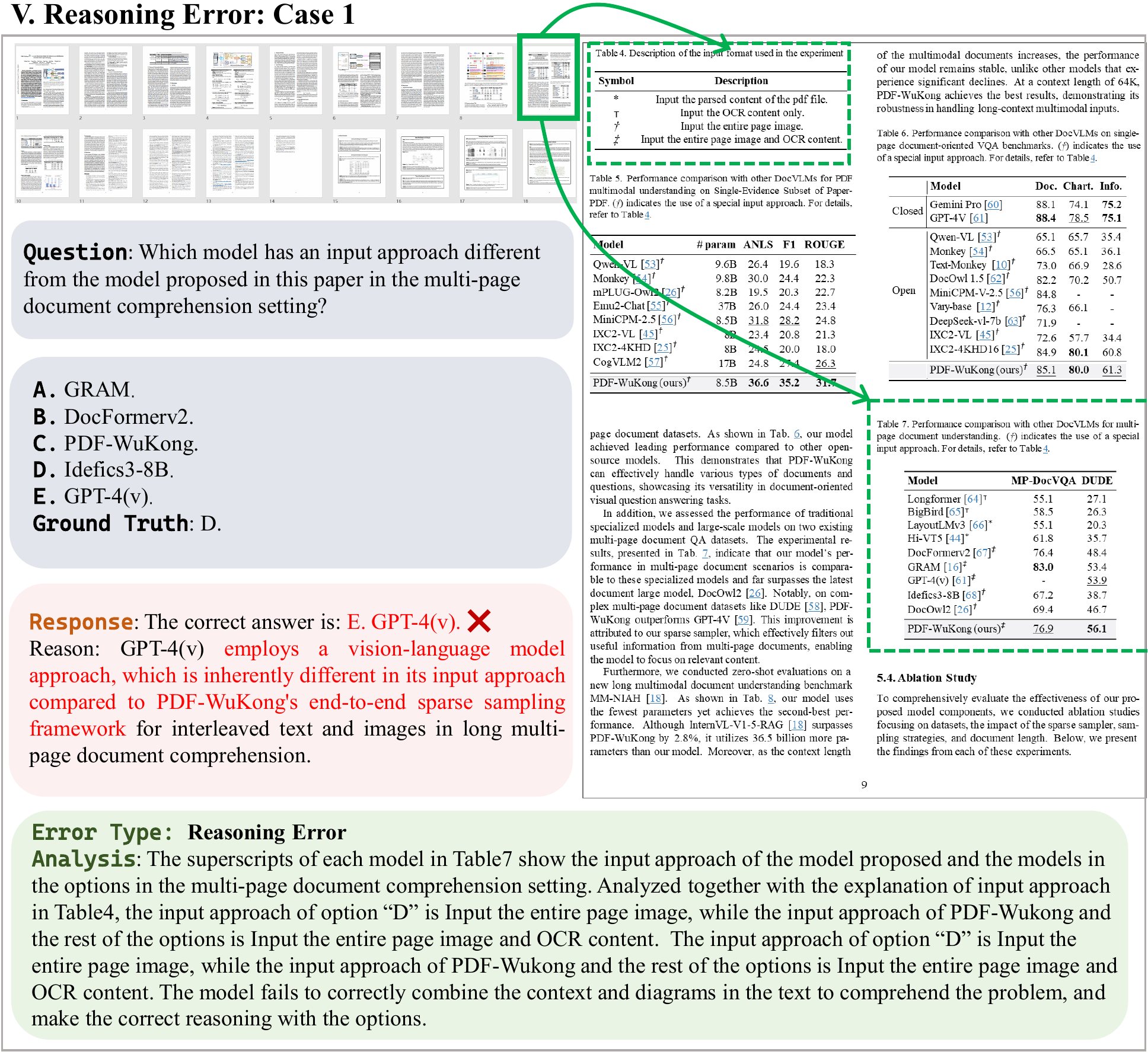}
    \caption{Illustration of a Reasoning Error Case. The figure demonstrates how the model fails to correctly interpret input approach differences between models despite clear evidence from Table 1 and Table 7, misidentifying GPT-4(v) as having a distinct approach from other models.}
    \label{fig:reasoning_error}
\end{figure*}

\begin{figure*}
    \centering
    \includegraphics[width=\linewidth]{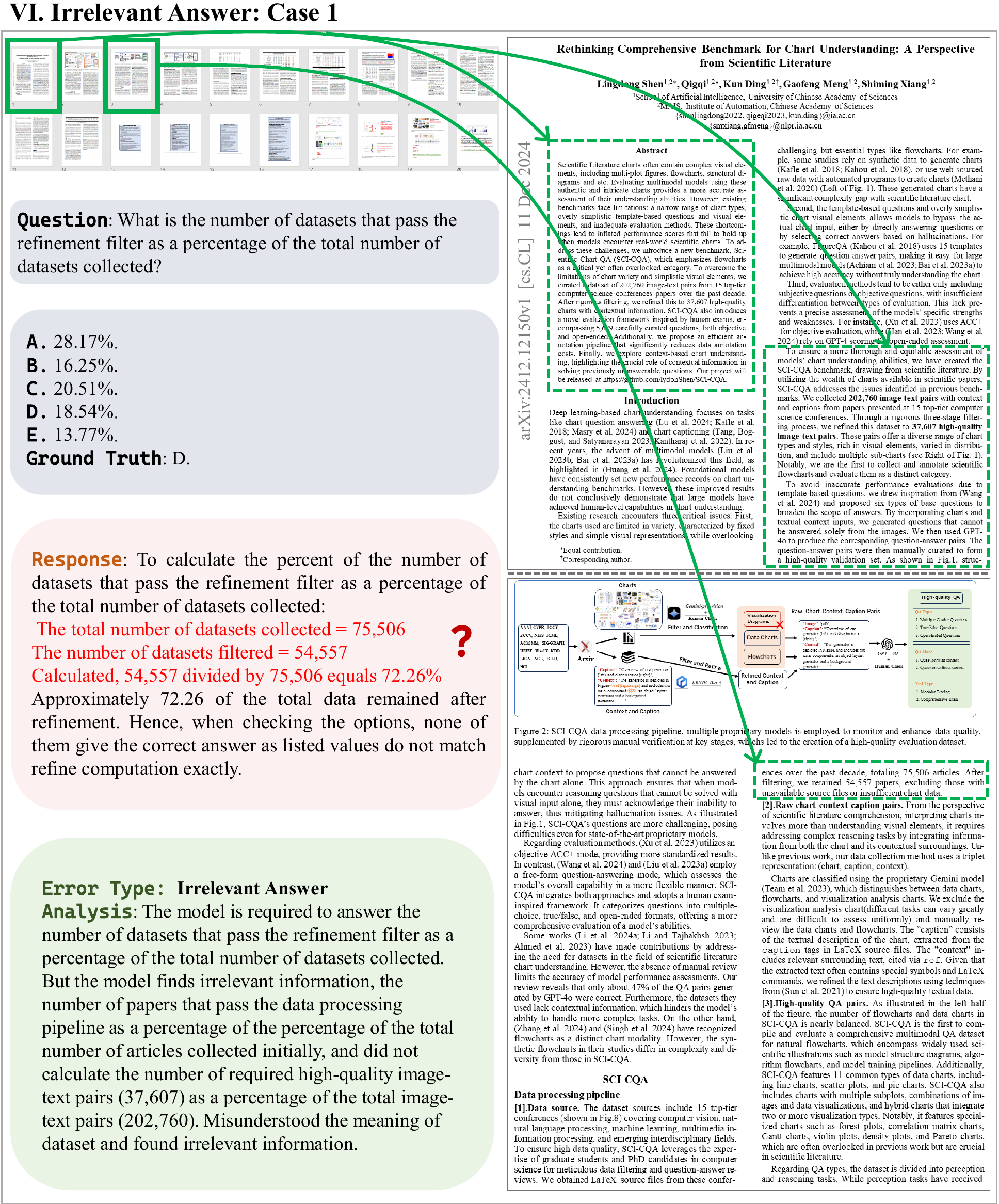}
    \caption{Illustration of an Irrelevant Answer Case. The figure demonstrates how the model misinterprets the question target, calculating the percentage of papers passing the pipeline (54,557/75,506) instead of the required percentage of high-quality image-text pairs (37,607/202,760), revealing a fundamental misunderstanding of the dataset definition}
    \label{fig:irrelevant_answer}
    
\end{figure*}

\begin{figure*}
    \centering
    \includegraphics[width=\linewidth]{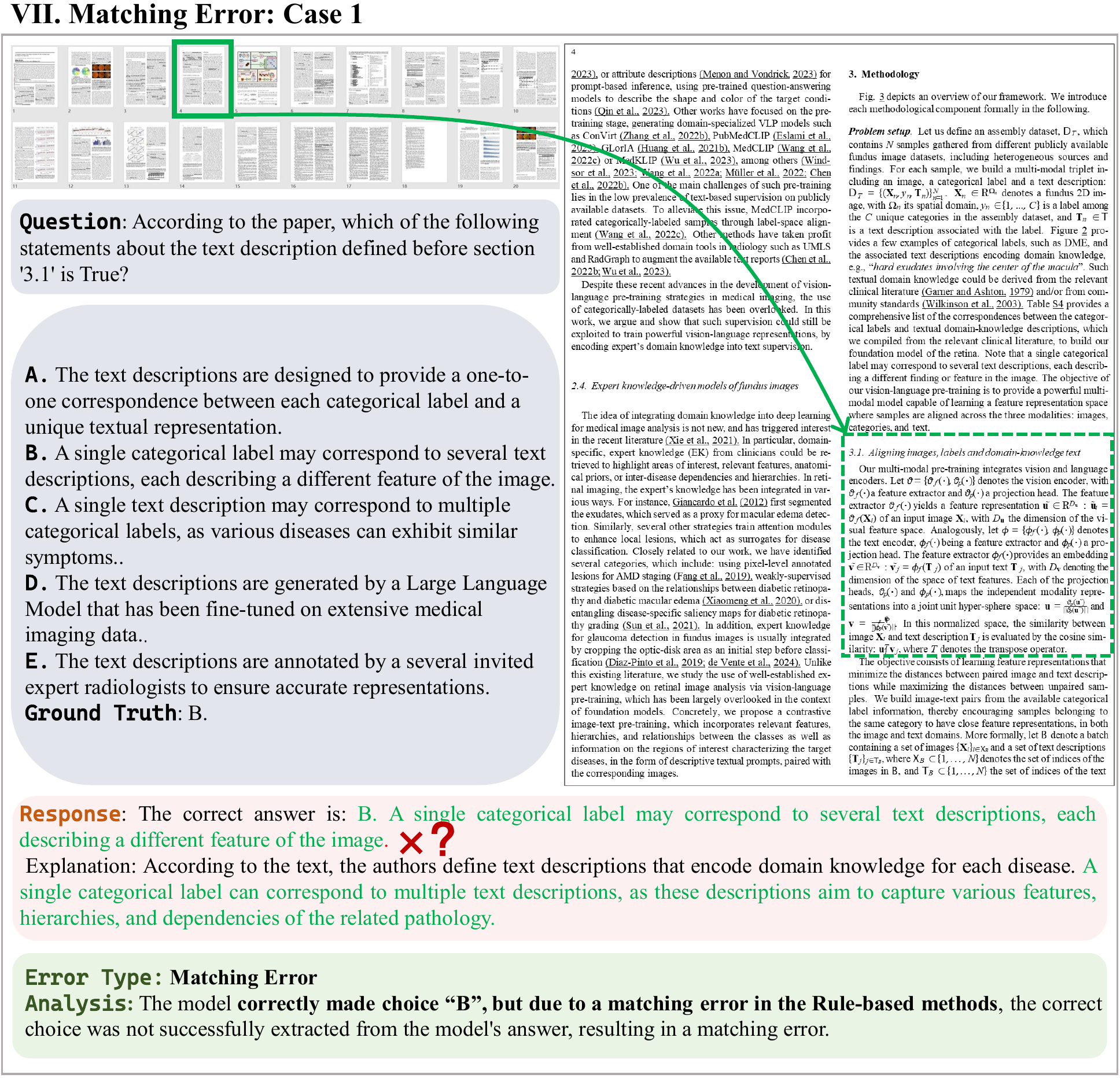}
    \caption{Illustration of a Matching Error Case. }
    \label{fig:matching_error}
\end{figure*}

We conducted systematic error analysis of GPT-4o's performance on our benchmark to investigate its limitations in cross-source reasoning within scientific papers. Through manual examination of 109 incorrect responses, we identified seven distinct error categories. A comprehensive analysis of all error categories, accompanied by representative examples, is presented in  (\cref{fig:hallucinated_evidence,fig:incomplete_evidence,fig:perceptual_error,fig:extractor_error,fig:reasoning_error,fig:irrelevant_answer,fig:matching_error}).

\subsection{Performance Across Evidence Types}
We analyze model performance across different evidence source types, with detailed results presented in \cref{fig:radar_chart_for_evidence_type}. The radar chart visualization demonstrates GPT-4o's consistent superiority across most categories compared to the other five VLMs. Particularly in text comprehension tasks, both QwenVL-2.5-72B and GPT-4o achieve notable accuracy (68.57 and 65.71 respectively), likely benefiting from their extensive pretraining corpora. 

However, substantial performance degradation is observed in cross-source integration tasks featured in MMCR, particularly in Figure-Text Comprehension, Figure-Table Comprehension, and Multi-Figure Comprehension, where the majority of VLMs achieve accuracy scores below 50. The pronounced disparity between single-source and cross-source task performance reveals a fundamental limitation: while MLLMs exhibit proficiency in individual modality processing, they demonstrate reduced effectiveness in tasks requiring synthesis of information from heterogeneous sources.

\subsection{Annotation requirements}
Before initiating the formal annotation process, a systematic taxonomy of task types and subject domains was established. This methodological framework ensures annotation consistency and maintains rigorous quality standards across the dataset construction process. Rigorous quality control protocols were implemented throughout the annotation process to establish a robust benchmark for evaluating the comprehensive capabilities of VLMs. Question formulation followed a structured protocol that integrates document-specific content with domain knowledge requirements, establishing a rigorous framework for in-depth assessment of scientific paper comprehension. The questions in MMCR are designed to evaluate comprehensive document understanding, specifically focusing on cross-source reasoning capabilities. The detailed evaluation requirements are illustrated in \cref{fig: Annotation requirements}.

\subsection{Annotation process}
The annotation process was conducted by expert annotators,  who underwent comprehensive training to ensure annotation consistency and quality standards. The standardized training protocol comprised several systematic phases: 1) \textbf{Initial standardization:} The project leader provided annotated sample papers to the annotators, which were subject to multiple rounds of verification. This iterative process ensured that the annotators fully understood the expectations and standards required for the annotation. 2) \textbf{Domain-specific allocation:} Annotators were assigned to subject domains aligned with their primary research expertise, selecting one to two domains from predefined categories. Within each domain, five representative publications were systematically identified for annotation. This domain-specific allocation ensures optimal alignment between expert knowledge and content analysis, maintaining annotation quality and disciplinary rigor. 3) \textbf{Quality assurance:} Completed annotations underwent systematic review by the project coordinator to ensure adherence to established protocols. When deviations from annotation standards were identified, annotators received structured feedback and supplementary training for remediation. This iterative quality assurance process continued until all annotations achieved compliance with predetermined quality benchmarks. 4)\textbf{Supplementary annotation:} In the final phase, expert annotators conducted supplementary annotation rounds in accordance with established protocols to expand the dataset while maintaining consistency standards.




\begin{figure}
\centering
\definecolor{brown}{RGB}{210,180,140}
\definecolor{grassgreen}{RGB}{112,173,71}
\begin{tcolorbox}[colback=lightgray!20,colframe=grassgreen, title= Examples for Annotation Requirements]
\textbf{Requirement I:} \\
``The question is specifically designed to examine the cross-source reasoning ability of the model in scientific papers, and it must be correctly derived exclusively from the designated information source, not from any additional information source. Annotators must strictly adhere to this requirement."

\textbf{Requirement II:} \\
``In order to mitigate the risk of the model relying exclusively on prior knowledge to generate responses, the annotated questions and answers were meticulously structured to ensure that accurate responses could only be derived by synthesizing information distributed across multiple pages of the document. This approach prevents the model from bypassing the intended reasoning process and reduces the potential for information leakage or unintended biases that might arise from relying on external knowledge."

\textbf{Requirement III:}\\
``To further ensure that the model genuinely comprehends the content of each paper, at least eight questions were annotated for every paper. This requirement was set to ensure a robust and comprehensive evaluation of the model's understanding across a variety of aspects within each paper."
\end{tcolorbox}

\caption{Annotation requirements}
\label{fig: Annotation requirements}
\end{figure}


\end{appendices}

\end{document}